\documentclass[10pt,final]{IEEEtran}

\usepackage{amsfonts}
\usepackage{extarrows}
\usepackage{enumerate}
\usepackage{amsmath}
\usepackage{amssymb}
\usepackage{makecell}
\usepackage{graphicx}
\usepackage{subfigure}
\usepackage{caption}
\usepackage{dsfont}
\usepackage{color}
\usepackage[square,comma,sort&compress,numbers]{natbib}

\usepackage{algorithmicx}
\usepackage[noend]{algpseudocode}
\usepackage{algorithm}

\usepackage{hyperref}
\hypersetup{
    colorlinks,
    citecolor=black,
    filecolor=black,
    linkcolor=black,
    urlcolor=black
}

\allowdisplaybreaks[4]

\usepackage{balance}

\begin{document}

\title{Design and Experimental Evaluation of a Hierarchical Controller for an Autonomous Ground Vehicle with Large Uncertainties}

\author{Juncheng Li, Maopeng Ran, and Lihua Xie, \emph{Fellow, IEEE}
\thanks{This work was supported by Delta-NTU Corporate Lab through the National Research Foundation Corporate Lab@University Scheme. \emph{(Juncheng Li and Maopeng Ran are co-first authors. Corresponding author: Lihua Xie.)}}
\thanks{J. Li, M. Ran, and L. Xie are with the School of Electrical and Electronic Engineering, Nanyang Technological University, Singapore 639798, Singapore (email: juncheng001@ntu.edu.sg; mpran@ntu.edu.sg; elhxie@ntu.edu.sg).}}

 \markboth{}
\markboth{}
\maketitle

\begin{abstract}

Autonomous ground vehicles (AGVs) are receiving increasing attention, and the motion planning and control problem for these vehicles has become a hot research topic. In real applications such as material handling, an AGV is subject to large uncertainties and its motion planning and control become challenging. In this paper, we investigate this problem by proposing a hierarchical control scheme, which is integrated by a model predictive control (MPC) based path planning and trajectory tracking control at the high level, and a reduced-order extended state observer (RESO) based dynamic control at the low level. The control at the high level consists of an MPC-based improved path planner, a velocity planner, and an MPC-based tracking controller. Both the path planning and trajectory tracking control problems are formulated under an MPC framework. The control at the low level employs the idea of active disturbance rejection control (ADRC). The uncertainties are estimated via a RESO and then compensated in the control in real time. We show that, for the first-order uncertain AGV dynamic model, the RESO-based control only needs to know the control direction.  Finally,  simulations and experiments on an AGV with different payloads are conducted.  The results illustrate that the proposed hierarchical control scheme achieves satisfactory motion planning and control performance with large uncertainties.

\end{abstract}


\IEEEpeerreviewmaketitle

\section{Introduction}

Nowadays, autonomous ground vehicles (AGVs) are playing an ever-increasing role in both civilian and military fields. These devices can increase productivity, decrease costs and human faults. However, in real applications (e.g., material handling in warehouses), an AGV is characterized by uncertain and challenging operational conditions, such as different payloads, varying ground conditions, and manufacturing imperfection \cite{R3,R14}. In this paper, we aim to solve the motion planning and control problem of AGVs with large uncertainties.

In general, there are three basic phases and modules in the AGV motion planning and control system, i.e., path planning, trajectory tracking control, and dynamic control \cite{R1}. The planning phase generates a feasible path for the AGV to follow. In the literature, various kinds of planning algorithms have been developed, such as A$^*$ algorithm \cite{A-algorithm}, D$^*$ algorithm \cite{D-algorithm}, and rapidly exploring random trees (RRT) \cite{RRT}. With a prior map of the environment, an AGV is able to plan a desired trajectory in real-time. The trajectory tracking control attempts to produce the velocity commands to enable the vehicle to track the planned path. So far, many approaches have been applied to AGV trajectory  tracking control, such as classical PID control \cite{R5}, sliding mode control \cite{R6}, robust control \cite{R7}, and intelligent control \cite{R3}.  However, realistic constraints imposed by the AGV model and physical limits cannot be effectively handled in theses methods.

In fact, the AGV path planning and trajectory tracking control with constraints can be naturally formulated into a constrained optimal control problem \cite{R8}. Therefore, model predictive control (MPC), which is capable of systematically handling constraints, became a well-known method to solve the AGV path planning and trajectory tracking control problem \cite{R9,R10,R11,Ale-2020,Cheng-2020}. In \cite{Ale-2020}, a linear parameter varying MPC (LPV-MPC) strategy was developed for an AGV to follow the trajectory computed by an offline nonlinear model predictive planner (NLMPP). In \cite{Cheng-2020}, an MPC-based trajectory tracking controller which is robust against the AGV parameter uncertainties was proposed. The controller in \cite{Cheng-2020} can be obtained by solving a set of linear matrix inequalities which are derived from the minimization of the worst-case infinite horizon quadratic objective function. Note that in \cite{R9,R10,R11,Ale-2020,Cheng-2020}, the trajectory tracking control is independent from the planning phase. However, a better integration between the path planning and trajectory tracking control would be helpful for enhancing the overall performance of an AGV.
These observations motivate us to develop a novel MPC-based     path   planning   and   trajectory   tracking control  scheme  for  AGVs.
The original rough path generated by a global planner is smoothed using MPC. A velocity planner is developed to assign the reference speed along the optimal path. Then an MPC-based trajectory tracking controller is designed to track the resulting trajectory. Since both the path planning and trajectory tracking control problems are solved using MPC-based methods, the AGV is expected to achieve better tracking performance.

\begin{figure*}
\begin{center}
  \includegraphics[width=0.9\textwidth,trim=10 0 5 400,clip]{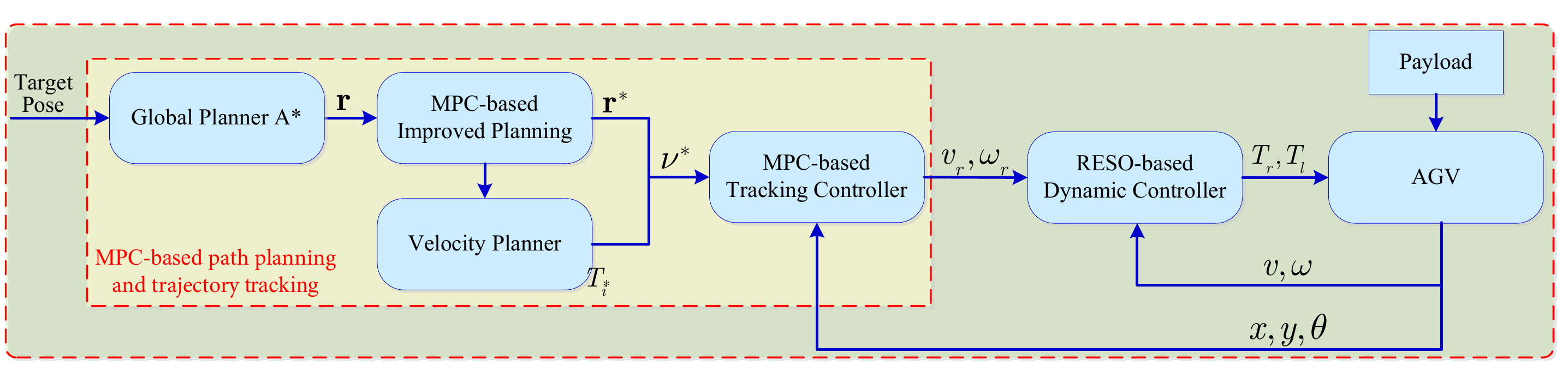}
  \caption{The proposed hierarchical control scheme for an AGV with large uncertainties.}
  \end{center}
\end{figure*}

The dynamic control, which performs in the AGV dynamic model level, aims to guarantee that the AGV moves according to the commands generated by the trajectory tracking controller. In practice, the AGV dynamic model is largely uncertain due to changing operational conditions. The active disturbance rejection control (ADRC), which is an efficient approach to handle uncertainties,  has received an increasing attention in recent years \cite{Han-2009,R15,R16}. The basic idea of ADRC is to view the uncertainties as an extended state of the system, and then estimate it using an extended state observer (ESO), and finally compensate it in the control action. In this paper, this idea is employed to design the AGV dynamic controller. A reduced-order ESO (RESO) is leveraged to estimate the uncertainties in the dynamic model of the AGV. On the theoretical side, we find that for general first-order uncertain nonlinear systems, the RESO-based ADRC controller only needs to know the sign of the control gain (i.e., the control direction). This is the weakest condition on the control gain compared with the existing ADRC literature (e.g., \cite{Ran1,Ran2,Ran3,Khalil-2008,Khalil-2015,Guo,Wu,Jiang}). More importantly, this feature of the RESO-based ADRC controller is vital to handle the possibly largely uncertain control gain in the  AGV dynamic model.

The proposed overall control scheme for an AGV with large uncertainties has a hierarchical structure (see, Fig. 1), i.e., an MPC-based path planning and trajectory tracking control in the high-level to plan the trajectory and steer the vehicle, while the RESO-based dynamic control in the low-level to track the velocity commands and handle the uncertainties.  To verify the effectiveness of the developed hierarchical controller, we implement it in an industrial AGV platform with different payloads. The main contributions of this paper are threefold.

\begin{enumerate}
  \item An MPC-based path planning and trajectory tracking control strategy is developed for industrial AGVs. Compared with the existing results, the developed strategy enables a better integration between the path planning and trajectory tracking control to enhance the overall tracking performance.
  \item A RESO-based dynamic controller is designed to handle the large uncertainties. It is proved that for a first-order uncertain nonlinear system, the RESO-based controller only needs to know the control direction. This is a new theoretical contribution, and makes the controller especially suitable for the AGV dynamic control, since in real applications such as material handling, the control gain of the AGV dynamic model is largely uncertain but its sign is fixed.
  \item Experiments of an AGV with different payloads are conducted  in a warehouse. The maximal payload in the experiments is about double the weight of the AGV itself. This indicates that large uncertainties exist in the AGV dynamics, and it is a very challenging scenario that has been rarely considered in the literature.  The experimental results show that the proposed hierarchical control scheme is easy for implementation, satisfactory in motion planning and trajectory tracking control, and effective in handling large uncertainties.
\end{enumerate}

The remainder of this paper is organized as follows. In Section II, the AGV system model and problem statement are presented. In Section III, the proposed MPC-based path planning and trajectory control approach is introduced.   Section IV gives the design and analysis of the RESO-based controller. Simulation and  experimental  results  are provided in Section V to demonstrate the effectiveness and superiority of the proposed  hierarchical control scheme. Finally, Section VI concludes the paper.

\section{System Model and  Problem  Statement}

\begin{figure}
\begin{center}
  \includegraphics[width=0.26\textwidth,trim=0 0 300 500,clip]{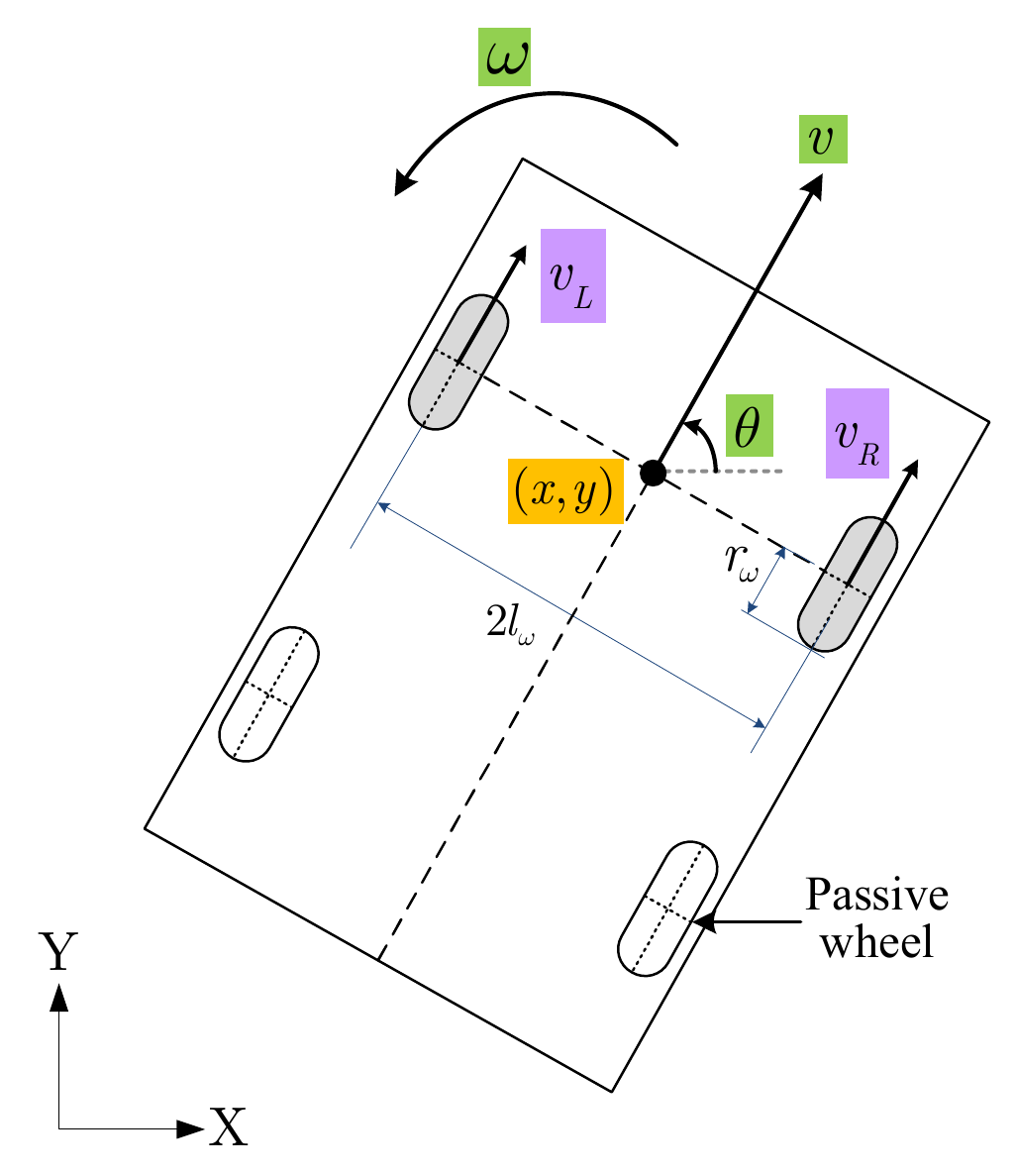}
  \caption{Schematic representation of the AGV.}
  \end{center}
\end{figure}

The schematic representation of the AGV is depicted in Fig. 2. The mathematical model of the AGV, including the kinematic equations and the dynamic equations, can be expressed as
\begin{equation}\label{eq1}
 \left\{\begin{aligned}
 \dot{x}(t)=& v(t)\cos(\theta(t)), \\
  \dot{y}(t)=& v(t)\sin(\theta(t)), \\
  \dot{\theta}(t)=& \omega(t), \\
  \dot{v}(t)=& \frac{T_r(t)+T_l(t)}{Mr_{\omega}}-\frac{f_{\rm{e}}(t)}{M}, \\
   \dot{\omega}(t)=& \frac{l_{\omega}(T_r(t)-T_l(t))}{2Ir_{\omega}}-\frac{\tau_{\rm{e}}(t)}{I},
 \end{aligned}\right.
\end{equation}
where  $[x(t),y(t)]^{\rm{T}}$ is the position of the AGV, $\theta(t)$ is the orientation, $[x(t), y(t), \theta(t)]^{\rm{T}}$ represents the pose of the AGV with respect to the coordinate $X-Y$;  $v(t)$ and $\omega(t)$ are the linear and angular velocities, respectively;  $M$ and $I$ are the mass and  moment of inertia of the AGV, respectively; $2l_{\omega}$ and $r_{\omega}$ are the distance between the two wheels and the radius of the wheel, respectively; $T_r(t)$ and $T_l(t)$ are the torques of the right and left wheels, respectively; $f_{\rm{e}}(t)$ and $\tau_{\rm{e}}(t)$ are the external disturbance force and torque, respectively.

Let ${v_L}$ and ${v_R}$ be the velocities of the left and right wheels of the AGV, respectively. Due to physical limitations, the velocities of the wheels are bounded by $\left| {v_L} \right| \leq v_{\max}$ and $\left| {v_R} \right| \leq v_{\max}$, where $v_{\max}$ is the maximal wheel speed. What is more, the linear and angular velocities of the AGV can be formulated as
\begin{subequations}
\label{equ:constraint}
\begin{align}
v &= ({v_L} + {v_R})/2, \\
w &= ({v_R} - {v_L})/({2l_w}).
\end{align}
\end{subequations}It follows from (\ref{equ:constraint}) that
\begin{equation}\label{equ:constraint3}
\left| v \right| + {l_w}\left| w \right| \le {v_{\max }}.
\end{equation}
The inequality \eqref{equ:constraint3} defines a diamond-shaped domain constraint on $v$ and $w$ \cite{R10}. Besides, the path planning and trajectory tracking control are inherently coupled \cite{R8}. Thus the first problem to be solved in this paper is to design an optimal path planning and trajectory tracking control approach, with the motion constraints systematically handled.

On the other hand, it can be observed from (1) that the dynamics of the AGV suffer from the external disturbance force and torque (such as friction force and torque). Most importantly, in some real applications, the mass and moment of inertia of the AGV platform are largely uncertain, due to different payloads and load  distributions. For example, in our experiments,  the maximal payload of the AGV reaches double of its own weight (i.e., the mass of the AGV platform varies between $M_0$ and $3M_0$, where $M_0$ is the own weight of the AGV). The second problem to be solved in this paper is to design a robust controller which is easy to implement and capable of handling the large uncertainties.

\section{MPC-based Trajectory Planning and Tracking Control}

A global path consists of a sequence of waypoints. Each waypoint can be represented by $r_i=[x_i, y_i]^{\rm{T}}$. If the total number of the waypoints is $N_r$, the path can be represented by a list  $\mathbf{r}= \{r_i\}_0^{N_r}$. A trajectory is a mapping between the time domain and space domain of the path. It can be represented by a list $\nu = \{r_i,T_i\}_0^{N_r}$, where $T_i$ is the reference time for the waypoint $r_i$. In this paper, we assume a navigation map is prebuilt by simultaneous localization and mapping (SLAM), and a rough original path is obtained by a geometric path planning algorithm (e.g., A* algorithm). First, this original path is assigned with a constant reference velocity to execute an MPC-based optimization process, which is for improving the smoothness of the original path with all the constraints systematically handled.  Then, based on the smoothed path, a reference velocity planner is developed to assign the AGV speed. Finally, the MPC-based trajectory tracking controller is designed.

\subsection{MPC-based Improved Path Planning}\label{path_planning}

The main idea of the MPC-based improved path planning is generating an optimal path by simulating a tracking process using MPC along the original path. The new path is smoother than the original path, and is kinematically feasible for the AGV to follow.

Since only the spatial information is considered in the global path planning stage, a trajectory with a constant velocity is first generated based on the global path $\mathbf{r}= \{r_i\}_0^{N_r}$. In our study, the resolution of the environment map is assumed to be high enough, so the distance between adjacent reference waypoints $r_{k-1}$ and $r_k$ is small. The time interval $\Delta t_k$ can be approximated by
\begin{equation}\label{equ:delta_t}
\Delta t_k = \frac{{\left\| {r_{k+1} - r_{k}} \right\|}}{{{v_{c }}}},
\end{equation}	
where the reference velocity of the robot at position $r_k$ is set to be a constant value ${v_{c}}$. Based on the time interval $\Delta t_k$, the corresponding reference time of each waypoint $r_k$ can be expressed as
\begin{equation}\label{equ:T}
T_k = \left\{
	\begin{aligned}
	& ~0, &&\mbox{if} \ k = 0  \\
	& \sum\limits_{i = 0}^{k-1} \Delta t_i, &&\mbox{if} \ {k} \in (0, N_r ]
	\end{aligned}\right.
\end{equation}

Then, an MPC-based tracking process is executed on the trajectory $\nu=\{r_i,T_i\}_0^{N_r}$ to generate a new path. The  optimal trajectory tracking control problem at time $T_k$ is formulated as
\begin{subequations}\label{equ:MPC}
	\begin{align}
	\mathop {\min }\limits_{{u_0},{u_1},...,{u_{H-1}}} &J_{\textrm{f}}(z_H,r_H)+\sum\limits_{i = 0}^{H-1} {
		J({z_i},{u_i},{r_i})
	}  \\
	\text{s.t.}\quad
	&{z_0} = z(T_k), \label{equ:MPC_inital} \\
	&{z_{i + 1}} = \chi({z_i},{u_i},\Delta t_{k+i}), \; \  i \in [0, H - 1]
	\label{equ:MPC_state_transition} \\
	&{z_i} \in \mathbb{Z}, \ {u_i} \label{equ:MPC_constraint} \in \mathbb{U}, \quad \quad \quad \ \; \; \forall i \geq 0
	\end{align}
\end{subequations}
where $J(\cdot)$ is the cost within the finite horizon $H$ and $J_\textrm{f}(\cdot)$ is the terminal cost. Here $z$ is the pose of the AGV, and $u$ is the control input which consists of the linear and angular velocities. This optimization problem is solved under several constraints. Eq. \eqref{equ:MPC_inital} represents the initial state constraint. The initial state $z_0$ is given by current state $z(T_k)$. Eq. \eqref{equ:MPC_state_transition} represents the kinematic constraint which is used to predict the motion of the AGV. According to the kinematic equation in  \eqref{eq1} and the reference trajectory, the kinematic constraint \eqref{equ:MPC_state_transition} can be specified by
\begin{equation}
z_{i+1} = z_i + \left[ {\begin{array}{*{20}{c}}
	{\cos ({\theta }_i)} & 0  \\
	{\sin ({\theta }_i)} & 0  \\
	0 & 1  \\
	\end{array}} \right]u_i\Delta t_{k+i}.
\label{equ:model_trajectory}
\end{equation}
Note that in \eqref{equ:model_trajectory}, the time interval $\Delta t_{k+i}$ should be small enough to reduce the discretization error. According to (4), it means that the trajectory should be adequately dense. Since the MPC predicts the AGV's future trajectory within a fixed horizon, a denser trajectory will lead to a higher computational complexity. Therefore, in practice, there exists a tradeoff between the model discretization accuracy and the computational complexity. Besides, Eq. \eqref{equ:MPC_constraint} represents the state and control input constraints, where $\mathbb{Z}$ and $\mathbb{U}$ are the feasible sets for $z$ and  $u$, respectively.

In the MPC framework, the cost function needs to be carefully defined to achieve satisfactory performance. For the path smoothing task, both tracking accuracy and smoothness of the control should be considered. Therefore, the complete cost function is designed as
\begin{equation}\label{equ:cost}
\sum\limits_{i = 1}^H {{{\widetilde{{z}}}_i^{\rm{T}}}} Q_i\widetilde{{z}}_i +\sum\limits_{i = 0}^{H-1} ({\widetilde{{u}}_i^{\rm{T}}}R_i\widetilde{{u_i}}+{\Delta u_i^{\rm{T}}}S_i{{\Delta {u_i}}}),
\end{equation}
where
\begin{subequations}
	\begin{align}
	&\widetilde{{z}}_i = r_{k+i} - {z_i},\\
	&\widetilde{{u}}_i = u_{k+i}^{\rm{ref}} - {u_i}, \\
	&\Delta {u}_i = {u_{i+1}} - {u_{i}},
	\end{align}
\end{subequations}
and $Q_i \in {\mathbb{R}^{3 \times 3}}$, $R_i \in {\mathbb{R}^{2 \times 2}}$, and $S_i \in {\mathbb{R}^{2 \times 2}}$ are positive definite weighting matrices. The cost function consists of three parts. $\widetilde{z}_i$ represents the distance between the state $z$ and the corresponding reference waypoint $r$ at time step $i$, so the first part of the cost function penalizes the distance between the smoothed path and original path within the time horizon $H$. $\widetilde{u}_i$ represents the difference between the control input $u$ and the reference velocity $u^{\rm{ref}}$ at time step $i$. Since a constant velocity $v_c$ is used to generate the trajectory, here $u^{\rm{ref}}=[v_c, 0]^{\rm{T}}$. $\Delta u_i$ represents the variation between two successive control inputs. The third part of the cost function penalizes the fluctuation of the control signal and makes the robot motion smoother. 

Note that the optimization problem \eqref{equ:MPC} is non-convex due to the inclusion of the  non-convex constraint \eqref{equ:MPC_state_transition}. In this study, we use the interior point method \cite{R17} to solve the  non-convex optimization problem \eqref{equ:MPC}. This method implements an interior point line search filter to find a local optimal solution. Since our objective is to find an optimal path close to the original path,  the original path is leveraged to initialize the optimization problem, which makes it very likely to converge to the global optimal solution.
The optimal predicted states $\bf{z}^*$ and control inputs $\bf{u}^*$ are denoted by
\begin{equation}\label{equ:uv}
{\ \; \bf{z}^*} = [ z_1^*, z_2^*, ..., z_H^*]^{\rm{T}}, \ \; {\bf{u}^*} =[u_0^*,u_1^*,...,u_{H - 1}^*]^{\rm{T}}.
\end{equation}
The predicted states $\bf{z}^*$ can be regarded as a new path which consists of $H$ waypoints. In the optimization problem, the tracking error and control input fluctuation have been penalized, so the new path should be both close to the original path and smooth enough. What is more, since the kinematic model and the velocity limits are considered during the optimization process, the new path is kinematically and physically feasible for the AGV.

If the prediction horizon $H$ is selected as the length of the original path $N_r$, the new optimal path $\mathbf{r}^*$ can be  directly obtained by $\bf{z}^*$. We denote this global optimal path as $\mathbf{r}^*_{N_r}$. However, in this case, the dimension (number of the decision variables) of the optimization problem is quite large and the computation will become time-consuming. Here, to solve this problem, we propose a piecewise path generation approach to achieve an efficient planning.

To begin with, a straightforward method is given. The MPC problem \eqref{equ:MPC} within a proper time horizion $H$ is solved. Then, based on the optimization result, $\mathbf{z}^*$ is extracted to the new path, and the final predicted state $z^*_H$ is used to update the initial constraint \eqref{equ:MPC_inital} which generates the next MPC problem. This process is repeated until the whole original path is replaced by the optimal predicted states. However, the path generated in this way maybe not smooth at the piecewise points. To overcome this limitation, we overlap the time horizon of the successive optimization process. Specifically, in each optimization step, only the first $H_{\rm{u}}$ elements of $\mathbf{z}^*$ is extracted to generate the new path and the initial state of the next MPC problem is set as $z^*_{H_{\rm{u}}}$, where $H_{\rm{u}}$ denotes the update horizon which satisfies $H_{\textrm{u}}<H$. In this way, the obtained path $\mathbf{r}^*$ is smoother than the original path and close to the global optimal path $\mathbf{r}^*_{N_r}$. The proposed MPC-based improved path planning is summarized in Algorithm \ref{alg:planning}.
\begin{algorithm}
	\caption{MPC-based improved path planning}
	\label{alg:planning}
	\begin{algorithmic}[1]
		\Require Original path $\mathbf{r}= \{r_i\}_0^{N_r}, r_i=[x_i,y_i]^{\rm{T}}$; Reference velocity $v_c$; Planning horizon $H$; Update horizon $H_{\textrm{u}}$
	    \Function {Simulated path improvement }{$\mathbf{r}$}
		\State Generate a trajectory $\mathbf{\nu} = \{r_i,T_i\}_0^{N_r}$ based on $v_c$
		\State Initial state $z_0 \gets r_0$ and time step $k \gets 0$
		\While {$k <N_r $}
		\State Initialization: $\{z_1,\ldots,z_H\} \gets \{r_{k+1},\ldots,r_{k+H}\}$
		\State Solve the MPC problem \eqref{equ:MPC}, obtain $\bf{z}^*$ and $\bf{u}^*$
		\State $\{r^*_k,\ldots,r^*_{k+H_{\textrm{u}}-1}\} \gets \{z^*_1,\ldots z^*_{H_{\textrm{u}}}\}$
		\State $k \gets k+H_{\textrm{u}}$
		\State $z_0 \gets z^*_{H_{\textrm{u}}}$
		\EndWhile
		\State \Return $\mathbf{r^*}= \{r^*_i\}_0^{N_r}, r_i^*=[x_i^*,y_i^*,\theta_i^*]^{\rm{T}}$
		\EndFunction
	\end{algorithmic}
\end{algorithm}

\subsection{Reference Velocity Planning}

In our approach, the trajectory generation is decomposed into spatial path generation and reference velocity (speed profile) planning. A reasonable and efficient reference velocity planning is necessary to obtain safe and comfortable navigation behavior. Based on the MPC-improved path, a velocity planner is developed in this subsection to assign the reference speed along the path, which completes the trajectory planning phase.

In real applications, the need for safe robot motion (e.g., to prevent rollover) should be considered. It is dangerous for an AGV to turn sharply at a high forward speed. However, the velocity constraint \eqref{equ:constraint3} cannot provide the guarantee of the safety. To solve this problem, we introduce a new parameter $c_v\geq 1$ into the constraint \eqref{equ:constraint3}, that is
\begin{equation}\label{equ:constraint_new}
\left| {v} \right| + {l_wc_v}\left| {w} \right| \le {v_{\max }}.
\end{equation}
By adjusting the parameter $c_v$, the velocity constraint \eqref{equ:constraint_new} can satisfy different requirements of the safety level. The linear constraint \eqref{equ:constraint_new} is utilized to generate the reference velocity (speed profile). Firstly, we approximate $w_i$ based on the path $\mathbf{r}^*=\{r_i^*\}_0^{N_r}$. Note that now each waypoint in the path is represented by $r_i^*=\left[x_i^*,y_i^*,\theta_i^*\right]^{\rm{T}}$, where $p_i^*=\left[x_i^*,y_i^*\right]^{\rm{T}}$ denotes the reference position, and $\theta_i^*$ denotes the reference heading angle of the AGV. Denote the distance between adjacent waypoints $r_i^*$ and $r_{i+1}^*$ as $d_i$, and the arc length from the origin to the waypoint $r_i^*$ as $s_i$. It can be computed that
\begin{equation}
    d_i =\left\| {p^*_{i+1}-p_i^*} \right\|, \
    s_i= \sum\limits_{k = 0}^{i-1} {d_k }.
\end{equation}
Since the time interval $\Delta t_i$ and the distance $d_i$ are small, the reference velocity and time derivative of the heading angle can be approximated by
\begin{equation}\label{viref}
    v_i^{\rm{ref}}={d_i \over \Delta t_i},
\end{equation}
\begin{equation}\label{dot_theta}
\dot \theta_i={\theta^*_{i+1}-\theta^*_i \over \Delta t_i}.
\end{equation}
According to \eqref{viref} and \eqref{dot_theta}, we  obtain
\begin{equation}
\dot \theta_i={\theta^*_{i+1}-\theta^*_i \over d_i} v_i^{\rm{ref}}.
\end{equation}

It can be observed from the AGV kinematic equations in \eqref{eq1} that the term $\dot \theta_i$ represents $w_i^{\rm{ref}}$ in the planning phase. However, if $\dot \theta_i$ is approximated by the discrete difference equation \eqref{dot_theta}, it may perform a discontinuous phenomenon and leads to an uncomfortable speed profile. To solve this problem, the polynomial curve fitting approach is leveraged to obtain a smooth reference velocity profile.  The path $\mathbf{r}^*$ can be parameterized using the
arc length along the path $s$, e.g., $\mathbf{r}^*(s_i)=r_i^*$. Similarly, the heading angle can be also parameterized by $\theta^*(s_i)=\theta_i^*$. A cubic polynomial curve fitting of heading angle $\theta^*$ with respect to $s$ is given by
\begin{equation}\label{equ:poly}
\theta^*(s)=c_3s^3+c_2s^2+c_1s+c_0,
\end{equation}
where $c_0$ to $c_3$ are coefficients determined by minimizing the fit error \cite{Reck-2000}. Generally, a single polynomial function cannot represent a long path precisely. Therefore, a piecewise polynomial curve fitting is implemented. Note that
\begin{equation}
w^{\rm{ref}} = {\frac{{\textrm{d}\theta^*}}{{\textrm{d}t}}}  = \frac{{\rm{d}\theta^* }}{{\textrm{d}s}}\frac{{\textrm{d}s}}{{\textrm{d}t}} = \frac{{\rm{d}\theta^* }}{{\textrm{d}s}}v^{\textrm{ref}}.
\end{equation}
As a result, the constraint \eqref{equ:constraint_new} can be rewritten as
\begin{equation}\label{speed_d_1}
\left| {{v^{\rm{ref}}}} \right| + {l_wc_v}\left| {\frac{{\rm{d}\theta^* }}{{\textrm{d}s}}}v^{\rm{ref}} \right| \le {v_{\max }}.
\end{equation}
For differential wheeled AGVs, it is assumed that $v^{\rm{ref}} \ge 0$, which means the AGV can only move forward. Consider the time-efficiency of the planned trajectory, the reference velocity $v^{\rm{ref}}$ is likely to be selected as large as possible. Therefore, the smoothed reference velocity at waypoint $r_i^*$ is specificed by
\begin{equation}\label{vref_con}
v_i^{\rm{ref}} = \frac{v_{\max }}{1 + \left| l_wc_v\left.\frac{\rm{d}\theta^* }{\textrm{d}s}\right|_{s=s_i} \right|}.
\end{equation}
This together with the MPC-improved path $\mathbf{r}^*$ yields the new trajectory $\nu^*=\{r_i^*, T_i^*\}_0^{N_r}$, where the timing information is calculated by
\begin{equation}\label{equ:final_T}
T^*_i = \sum\limits_{k = 0}^{i-1} {\frac{{\left\| {p^*_{k+1} - p^*_k} \right\|}}{{{v_i^{\rm{ref}}}}}}, \ i \in (0,N_r].
\end{equation}

Note that in this paper, the spatial and  temporal planning of the trajectory are separated. More specifically, the  path is first smoothed by optimization using a constant velocity $v_c$, and then the smoothed path is leveraged for the velocity planning. Our simulation and experimental experiences indicate that such a simple two-stage procedure suffices to achieve a satisfactory performance. In some cases, an iterative procedure between the path smoothing and velocity planning could further improve the performance, but with longer computation time.

\subsection{MPC-based Kinematic Controller}

In the online kinematic control stage, due to some practical issues such as localization error, the  control  sequences  generated  in  the trajectory  planning  stage cannot be directly  applied.  A trajectory tracking controller is needed to make the vehicle track the planned trajectory. What is more, those unnecessary aggressive maneuvers should also be avoided. Due to these considerations, the MPC-based trajectory tracking problem is formulated as
\begin{subequations}\label{equ:MPC_new}
	\begin{align}
	\centering
		\mathop {\min }\limits_{{u_0},\cdots,{u_{H-1}}}& \sum\limits_{i = 1}^H {{{\widetilde{{z}}}_i^{\rm{T}}}} Q_i\widetilde{{z}}_i +\sum\limits_{i = 0}^{H-1} ({\widetilde{{u}}_i^{\rm{T}}}R_i\widetilde{{u_i}}+{\Delta u_i^{\rm{T}}}S_i{{\Delta {u_i}}})\\
	&{z_0} = z_c(t), \label{equ:MPC_inital_new} \\
	&{z_1} = \chi(z_0,u_0,T^*_{k+1}-t), \label{equ:state_transition_01} \\
    &{z_{i + 1}} =\chi({z_i},{u_i},\Delta t_{k+i}^*), \; \  i \in [0, H - 1]
	\label{equ:MPC_state_transition_new} \\
	&{z_i} \in \mathbb{Z}, \ {u_i} \label{equ:MPC_constraint_new} \in \mathbb{U}, \quad \quad \quad \ \; \; \forall i \ge 0
	\end{align}
\end{subequations}
where $\widetilde{{z}}_i = r_{k+i}^* - {z_i}$, $\widetilde{{u}}_i = u_{k+i}^{\rm{ref}} - {u_i}$ and $\Delta {u_i} = {u_{i+1}} - {u_{i}}$. The optimal control problem needs to be solved at each sampling time. At time $t$, the localization system gives the estimated state of the robot $z_c(t)$, which is used to update the initial constraint \eqref{equ:MPC_inital_new}. The index $k$ in the kinematic constraint \eqref{equ:state_transition_01} and \eqref{equ:MPC_state_transition_new} satisfies $T^*_k\leq t<T^*_{k+1}$. By solving \eqref{equ:MPC_new}, the optimal predicted states $\mathbf{z}^*$ and inputs $\mathbf{u}^*$ are obtained, then only the first element of $\mathbf{u}^*$ is  applied to the AGV as the kinematic control input. At the next sampling time, the optimization problem \eqref{equ:MPC_new} is rebuilt and solved again. The whole process is repeated until the goal position is reached.

\section{RESO-based Dynamic Controller}

The kinematic controller generates the commands of the linear and angular velocities (denoted by $v_r$  and $\omega_r$, respectively) for the AGV to track the planned trajectory. The objective of the dynamic controller is to make the actual AGV velocities $v$ and $\omega$ follow $v_r$ and $\omega_r$, respectively. What is more, the large uncertainties (maybe introduced by friction, payloads, etc) in the dynamic equations of the AGV need to be properly handled. To accomplish this goal, we first design a RESO-based tracking controller for a class of general first-order uncertain nonlinear systems. Rigorous theoretical analysis is given to show that, the tracking error can be made arbitrary small under large uncertainties. This approach is then applied to design the dynamic controller for the AGV.

\subsection{RESO-based Tracking Control for First-Order Uncertain Nonlinear Systems}

Consider the following first-order uncertain nonlinear system
\begin{equation}\label{eq2}
\dot{\eta}(t)= f(\eta(t), \varpi(t))+b(\eta(t), \varpi(t))u(t),  ~ t\geq 0,
\end{equation}
where $\eta(t)\in\mathbb{R}$ is the state, $\varpi(t)\in \mathbb{R}$ is the  external disturbance, $f(\cdot), b(\cdot): \mathbb{R}\times \mathbb{R}\rightarrow \mathbb{R}$ are uncertain continuously differentiable functions. The objective of the control is to guarantee that the state of the system (\ref{eq2}), $\eta(t)$,  tracks a given reference signal $\varrho(t)$.

\emph{Assumption A1:} The external disturbance $\varpi(t)$ and its first derivative $\dot{\varpi}(t)$ are  bounded.

\emph{Assumption A2:} The reference signal $\varrho(t)$ and its first two derivatives  are bounded.

\emph{Assumption A3:} For all  $(\eta(t),\varpi(t))\in\mathbb{R}\times\mathbb{R}$, the control gain  $b(\cdot)\neq 0$, and its sign is fixed and known.

Note that large uncertainties exist in system (\ref{eq2}), since the drift dynamics $f(\eta(t),\varpi(t))$ is totally unknown, and only the sign of the control gain is known. Let $b_0(\eta(t)): \mathbb{R}\rightarrow \mathbb{R}$ be a nominal function of the control gain $b(\eta(t),\varpi(t))$.  In this paper, we only require the signs of $b(\cdot)$ and $b_0(\cdot)$ are the same. By Assumption A3 and without loss of generality, it is assumed that $b(\cdot),b_0(\cdot)>0$.  For system (\ref{eq2}), the total uncertainty is defined as
\begin{equation}\label{eq3}
\xi(t)=f(\eta(t),\varpi(t))+(b(\eta(t),\varpi(t))-b_0(\eta(t)))u(t).
\end{equation}
According to (\ref{eq2}) and (\ref{eq3}),  a first-order RESO is designed as
\begin{equation}\label{eq4}
\dot{\varsigma}(t)=\frac{1}{\varepsilon}L(\eta(t)-\varsigma(t))+b_0(\eta(t))u(t), \end{equation}
where $\varsigma(t)\in \mathbb{R}$  is the observer state, $\varepsilon<1$ is a small positive constant, and $L>0$ is the observer gain. The output of the RESO (\ref{eq4}) is
\begin{equation}\label{eq5}
\hat{\xi}(t)=\frac{1}{\varepsilon}L(\eta(t)-\varsigma(t)),
\end{equation}
which, is the estimate of the total uncertainty $\xi(t)$.

Based on the output of the RESO (\ref{eq4}), the control is designed as
\begin{equation}\label{eq6}
u(t)=\frac{K(\eta(t)-\varrho(t))-\hat{\xi}(t)+\dot{\varrho}(t)}{b_0(\eta(t))}\triangleq\psi(\eta(t),\varrho(t),\hat{\xi}(t)),
\end{equation}
where $K<0$. Furthermore,  inherited from the previous high-gain observer results \cite{Khalil-2015,Khalil-2008,Ran1,Ran2}, to protect the system from the peaking caused by the initial error and high-gain, the control to be injected into the system is modified as
\begin{equation}\label{eq7}
u(t)=M_u\textrm{sat}_{\varepsilon}\left(\frac{\psi(\eta(t),\varrho(t),\hat{\xi}(t))}{M_u}\right),
\end{equation}
 where $M_u$ is the bound selected such that the saturation will not be invoked under state feedback \cite{Khalil-2015,Khalil-2008}, i.e.,
 \begin{equation}\label{bound}
  M_u>\sup_{t\in[0,\infty)}\left|\frac{K(\eta(t)-\varrho(t))-f(\eta(t),\varpi(t))+\dot{\varrho}(t)}{b(\eta(t),\varpi(t))}\right|. \end{equation}
 The function $\textrm{sat}_{\varepsilon}(\cdot): \mathbb{R}\rightarrow \mathbb{R}$ is odd and defined by \cite{Khalil-2008}
\begin{equation*}
\textrm{sat}_{\varepsilon}(\ell)= \left\{
  \begin{aligned}
       &   \ell && \textrm{for} ~0\leq \ell \leq 1 \\
       &   \ell+\frac{\ell-1}{\varepsilon}-\frac{\ell^2-1}{2\varepsilon} && \textrm{for} ~1\leq \ell \leq 1+\varepsilon \\
      &          1+\frac{\varepsilon}{2} &&  \textrm{for} ~\ell>1+\varepsilon
        \end{aligned} \right.
\end{equation*}
It can be observed that $\textrm{sat}_{\varepsilon}(\cdot)$ is nondecreasing, continuously differentiable with locally Lipschitz derivative, and satisfies $\left|\frac{\textrm{dsat}_{\varepsilon}(\ell)}{\textrm{d}\ell}\right|\leq 1$ and $\left|\textrm{sat}_{\varepsilon}(\ell)-\textrm{sat}(\ell)\right|\leq \frac{\varepsilon}{2}$, $\forall \ell\in\mathbb{R}$, where $\textrm{sat}(\cdot)$ is the standard unity saturation function denoted by $\textrm{sat}(\ell)=\textrm{sign}(\ell)\cdot\min\{1,|\ell|\}$.

\emph{Theorem 1:} Consider the closed-loop system formed of (\ref{eq2}), (\ref{eq4}), (\ref{eq5}), and (\ref{eq7}). Suppose Assumptions A1 to A3 are satisfied, and the initial conditions $\eta(0)$ and $\varsigma(0)$ are bounded. Then for any $\sigma>0$, there exists $\varepsilon^{\dag}>0$ such that for any $\varepsilon\in(0,\varepsilon^{\dag})$,
\begin{equation}\label{eq8}
 |\xi(t)-\hat{\xi}(t)|\leq \sigma,  ~t\in [T, \infty), ~\forall T> 0,
\end{equation}
and
\begin{equation}\label{eq9}
 \lim_{t\rightarrow \infty} |\eta(t)-\varrho(t)|\leq \sigma.
\end{equation}

\emph{Proof:} See the Appendix. \IEEEQED

\emph{Remark 1:} It is now well-realized that the ESO-based control (or ADRC) needs a good prior estimate of the control gain \cite{Ran1,Ran2,Khalil-2008,Khalil-2015,Guo,Wu,Jiang}. In other words, the nominal control gain $b_0(\eta(t))$ should be ``close'' to the actual control gain $b(\eta(t),\varpi(t))$. However, the results in Theorem 1 show that for the first-order uncertain nonlinear system (\ref{eq2}), the RESO-based controller only relies on the knowledge of the control direction (i.e., the sign of $b(\eta(t),\varpi(t))$).  This result is very meaningful, since in practice it is much easier to obtain the knowledge of the control direction than an accurate prior estimate of the control gain.  In \cite{Xue2}, a similar conclusion was achieved via frequency-domain analysis for first-order linear uncertain systems. As far as the author's knowledge goes, this paper is the first work that conducts rigorous time-domain analysis for RESO-based control for first-order uncertain nonlinear systems.

\subsection{Dynamic Controller for an AGV with Large Uncertainties}

For the AGV dynamic controller design, we first consider the following transformation:
\begin{equation}\label{eq16}
    u_{v}(t)=T_r(t)+T_l(t), ~u_{\omega}(t)=T_r(t)-T_l(t).
\end{equation}
Then the AGV dynamic model can be rewritten as
\begin{equation}\label{eq17}
 \left\{\begin{aligned}
  \dot{v}(t)=&\frac{1}{Mr_{\omega}}u_v(t)-\frac{f_{\rm{e}}(t)}{M}, \\
  \dot{\omega}(t)=& \frac{l_{\omega}}{2Ir_{\omega}}u_{\omega}(t)-\frac{\tau_{\rm{e}}(t)}{I}.
 \end{aligned}\right.
\end{equation}
Note that in (\ref{eq17}), the control gains for the AGV linear and angular velocities (i.e., $\frac{1}{Mr_{\omega}}$ and $\frac{l_{\omega}}{2Ir_{\omega}}$) are largely uncertain due to different payloads, but their signs are fixed and known to the designer. Therefore, the developed RESO-based controller is capable of handling the large uncertainties in the AGV dynamic model.

By Theorem 1, the observers and controllers are designed as
\begin{equation}\label{eq18}
\left\{
\begin{aligned}
\dot{\varsigma}_v(t)=&\frac{1}{\varepsilon}L(v(t)-\varsigma_v(t))+b^v_0u_v(t),  \\
\hat{\xi}_v(t)=&\frac{1}{\varepsilon}L(v(t)-\varsigma_v(t)),
\end{aligned}
\right.
\end{equation}
\begin{equation}\label{eq19}
\left\{
\begin{aligned}
\dot{\varsigma}_{\omega}(t)=&\frac{1}{\varepsilon}L({\omega}(t)-\varsigma_{\omega}(t))+b^{\omega}_0u_\omega(t),  \\
\hat{\xi}_{\omega}(t)=&\frac{1}{\varepsilon}L(\omega(t)-\varsigma_{\omega}(t)),
\end{aligned}
\right.
\end{equation}
\begin{equation}\label{eq20}
u_v(t)=M_v\textrm{sat}_{\varepsilon}\left(\frac{K_v(v(t)-v_r(t))-\hat{\xi}_v(t)+\dot{v}_r(t)}{M_{v}b^v_{0}}\right),
\end{equation}
\begin{equation}\label{eq21}
u_{\omega}(t)=M_{\omega}\textrm{sat}_{\varepsilon}\left(\frac{K_{\omega}(\omega(t)-{\omega}_r(t))-\hat{\xi}_{\omega}(t)+\dot{\omega}_r(t)}{M_{\omega}b^{\omega}_{0}}\right),
\end{equation}
where $b_0^v$ and $b_0^{\omega}$ are the nominal values of $b^v=\frac{1}{Mr_{\omega}}$ and $b^{\omega}=\frac{l_{\omega}}{2Ir_{\omega}}$, respectively; $\hat{d}_v(t)$ and $\hat{d}_{\omega}(t)$ are the estimates of the uncertainties $d_v(t)\triangleq-\frac{f_{\rm{e}}(t)}{M}+(b^v-b^v_0)u_v(t)$ and $d_{\omega}(t)\triangleq-\frac{\tau_{\rm{e}}(t)}{I}+(b^{\omega}-b^{\omega}_0)u_{\omega}(t)$, respectively; $K_v$, $K_{\omega}<0$; $M_v$ and $M_{\omega}$ are the saturation bounds.

Finally, by (\ref{eq16}), the actual control commands are obtained by
\begin{equation}\label{eq22}
    T_r(t)=\frac{1}{2}(u_v(t)+u_{\omega}(t)), ~T_{l}(t)=\frac{1}{2}(u_v(t)-u_{\omega}(t)).
\end{equation}

\begin{figure}
    \centering
    \includegraphics[width=0.35\textwidth]{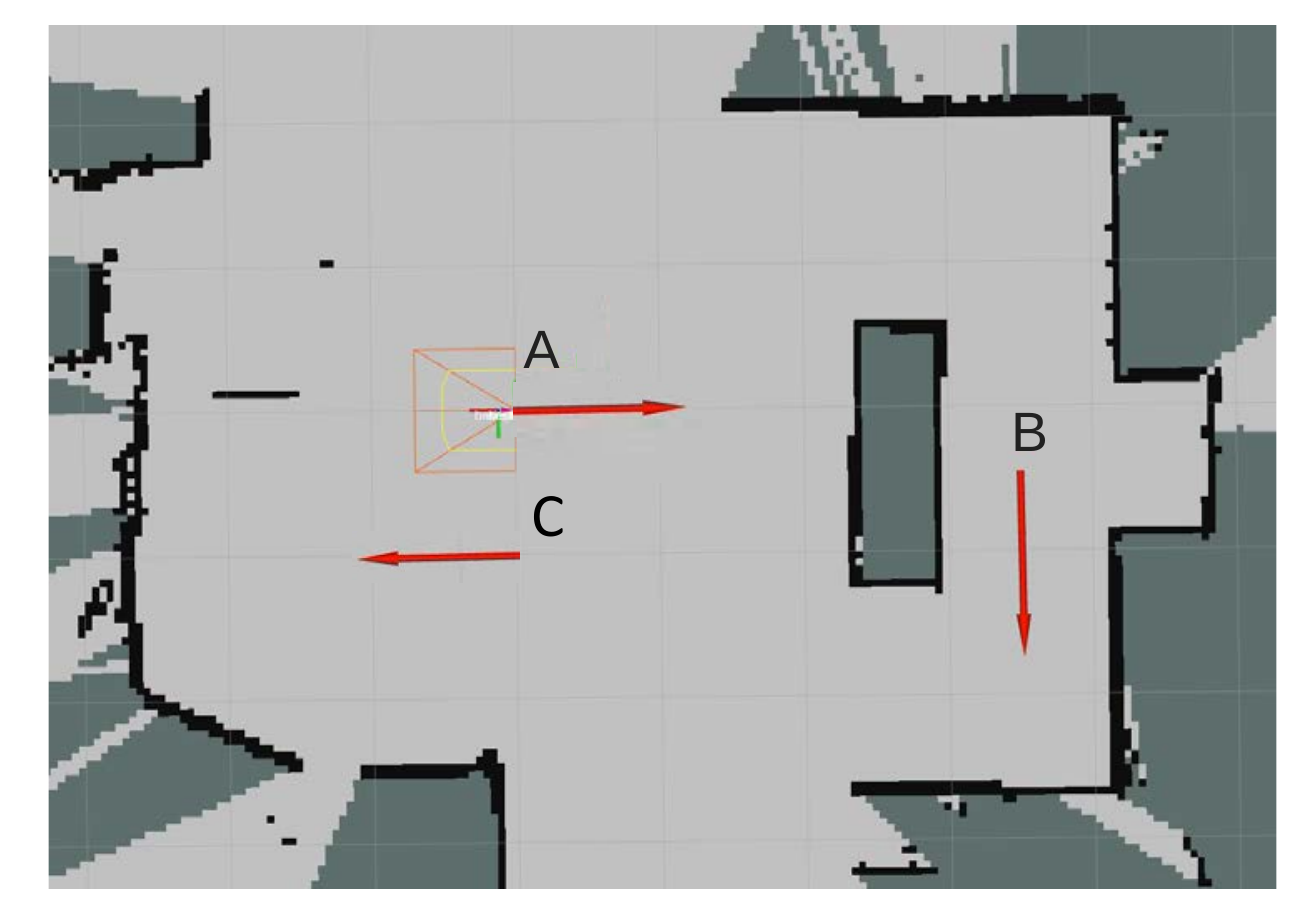}
    \caption{Simulation scenario. The AGV moves from point A to point B, and then comes back to point C. The red arrows represent the orientations of the AGV at the three points.}
    \label{Fig_sim_scenario}
\end{figure}

\begin{figure}
\begin{center}
 \includegraphics[width=0.38\textwidth]{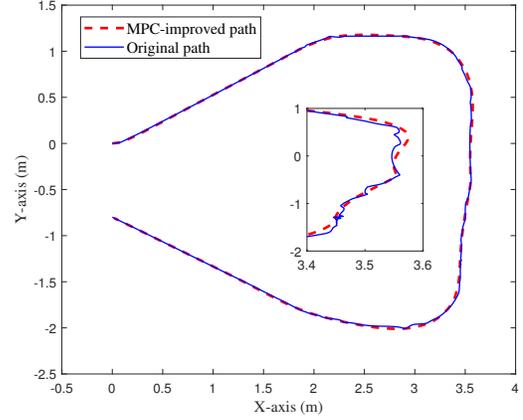}
 \caption{The original and MPC-improved paths.}
 \label{Fig_MPC_path}
 \end{center}
\end{figure}

\begin{figure}
\begin{center}
 \includegraphics[width=0.4\textwidth]{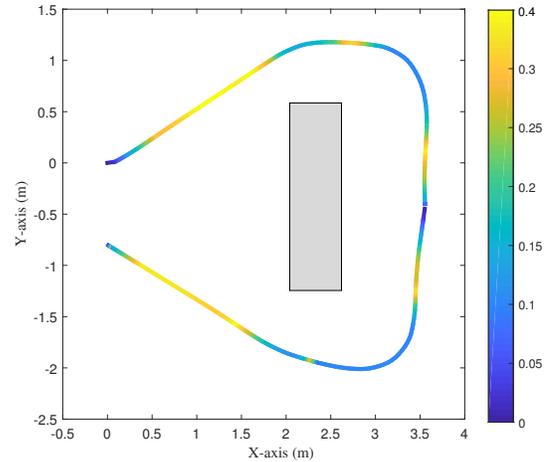}
 \caption{Velocity profile along the planned path. }
 \label{Fig_sim_tra}
 \end{center}
\end{figure}

\begin{figure}[!t]
\begin{center}
  \includegraphics[width=0.38\textwidth]{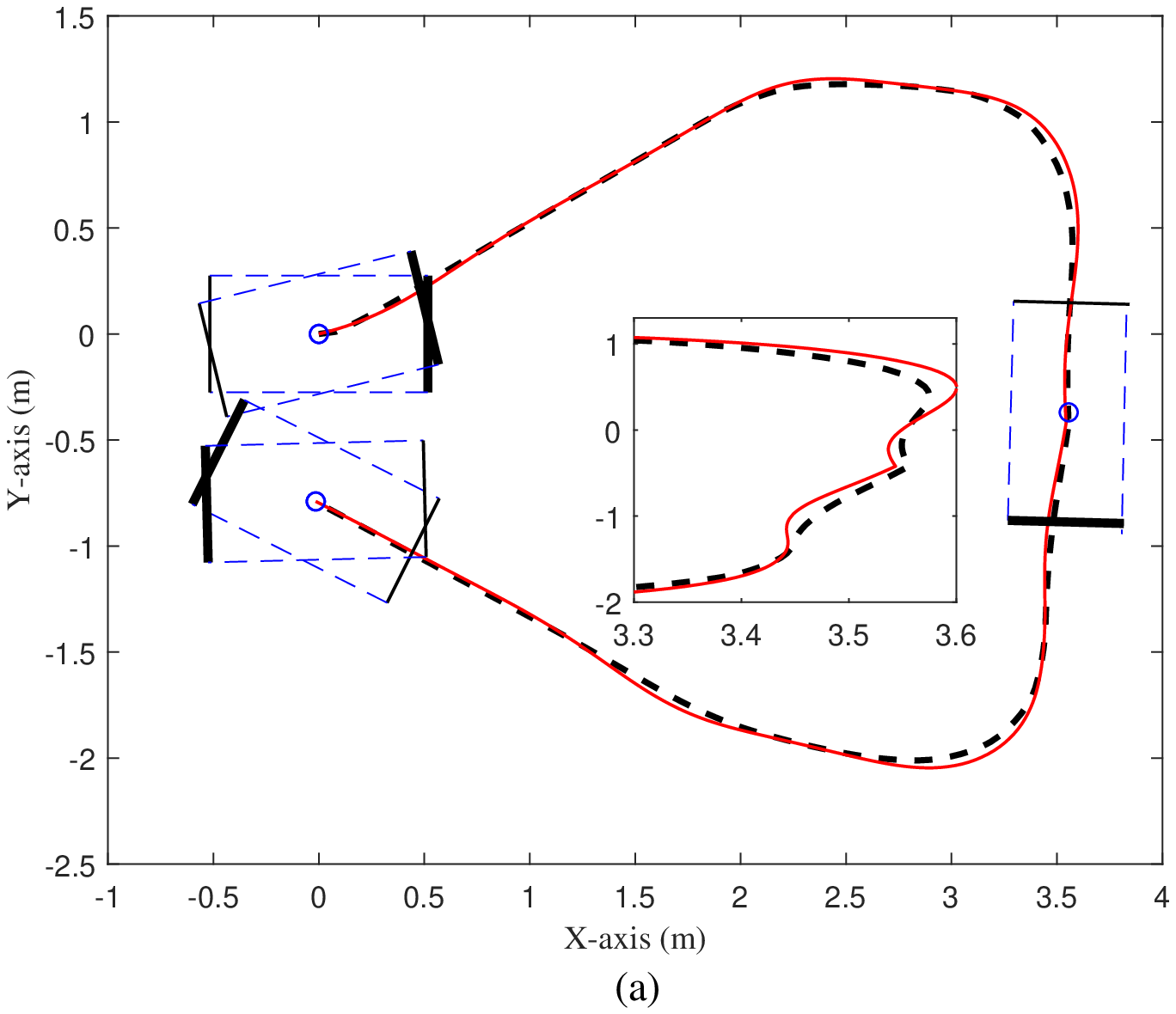}
  \includegraphics[width=0.38\textwidth]{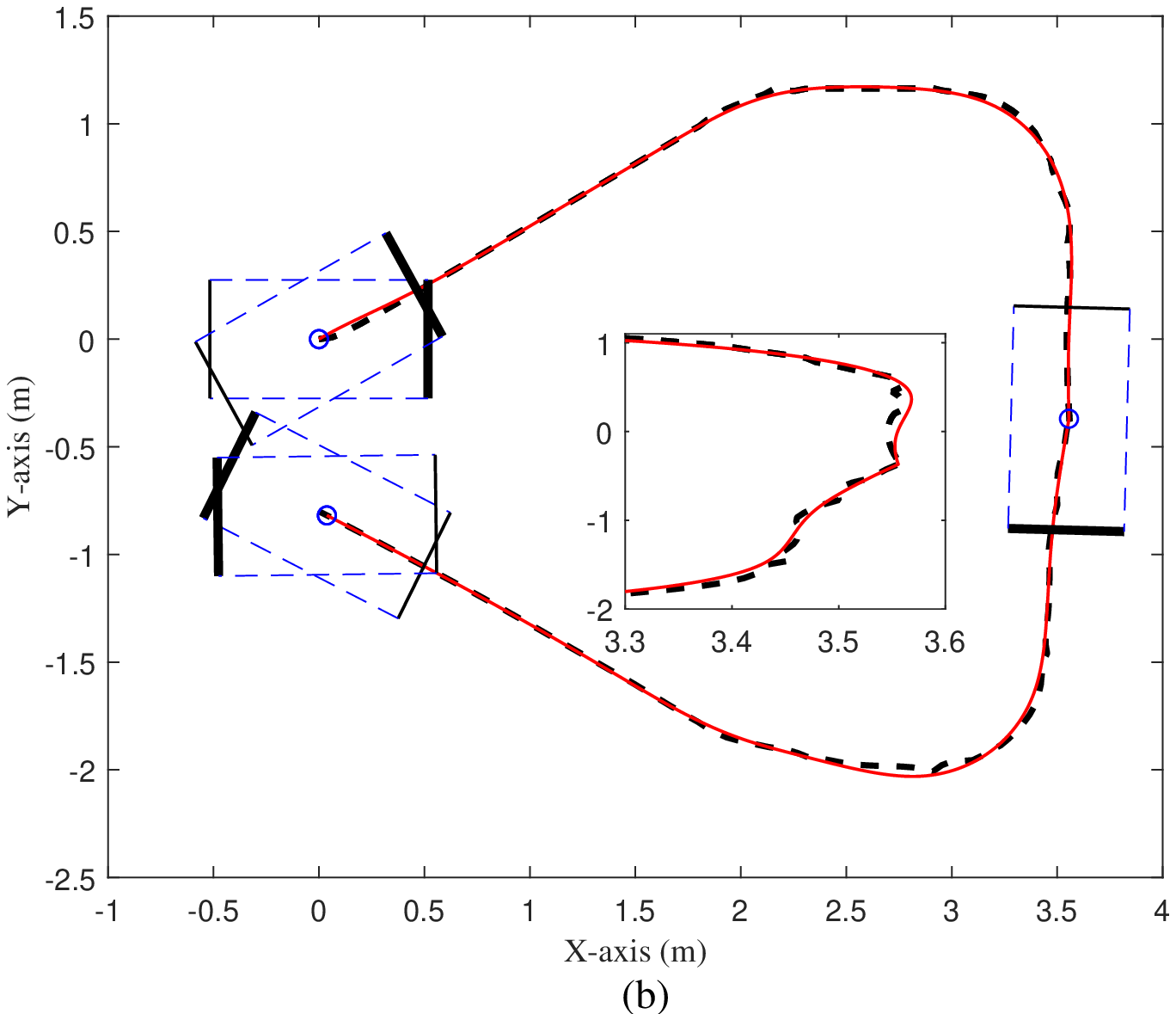}
  \includegraphics[width=0.38\textwidth]{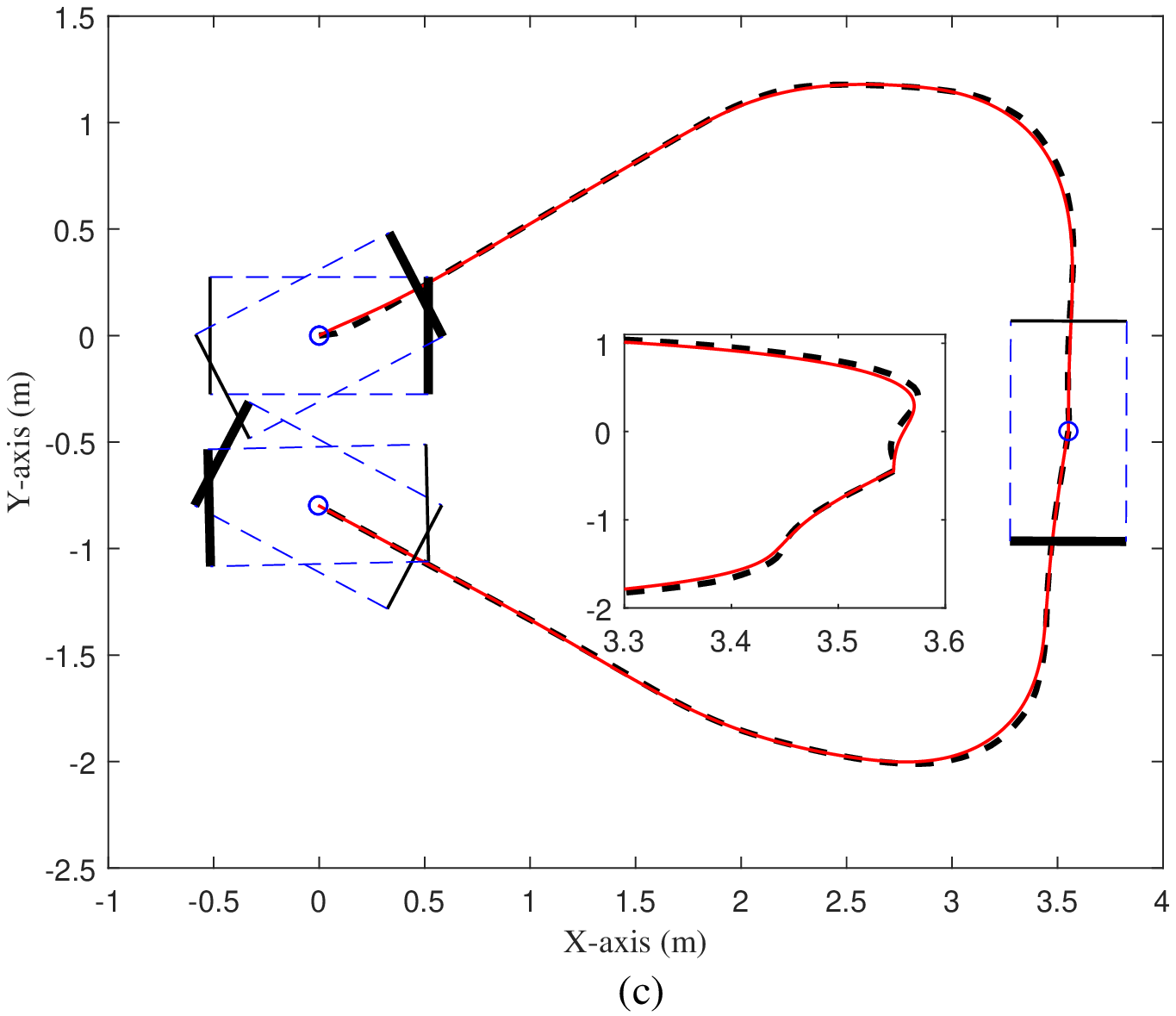}
  \end{center}
  \caption{Planned (black dashed lines) and actual trajectories (red solid lines) of different planning and tracking control schemes: (a) MPC-based trajectory  planning  +  PID  tracking  control; (b)  Global planner  A$^*$  +  MPC-based  tracking  control; (c) MPC-based trajectory planning and tracking control.}
 \end{figure}

 \begin{figure}[!t]
\begin{center}
  \includegraphics[width=0.38\textwidth]{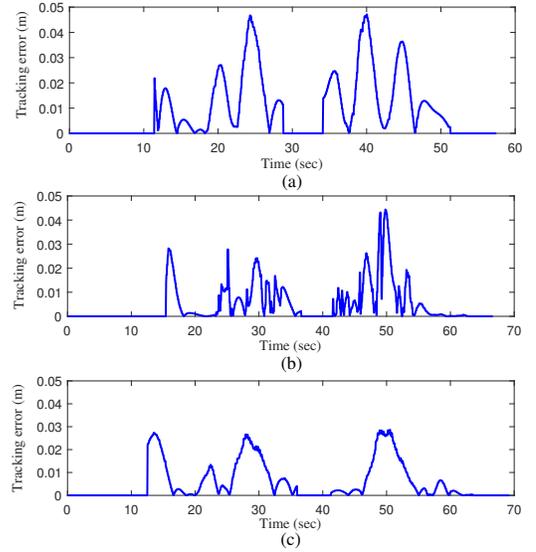}
 \caption{Tracking errors: (a) MPC-based trajectory planning  +  PID  tracking  control; (b)  Global planner  A$^*$  +  MPC-based  tracking  control; (c) MPC-based trajectory planning and tracking control. }
  \end{center}
 \end{figure}

\section{Simulation and Experimental Evaluation}

This section presents the simulation and experimental results of the proposed hierarchical control scheme. The MPC-based trajectory planning and tracking control algorithm is first implemented on \emph{Robot Operating System (ROS)} to tune the parameters, and to verify its superiority. The effectiveness of the RESO-based dynamic controller in handling large uncertainties is then tested in \emph{Matlab/Simulink}. Finally, experiments are conducted on an AGV in a warehouse environment with different payloads.

\subsection{Simulation Results}

The \emph{stage} simulator is used to build the simulation environment in \emph{ROS}. An  efficient optimization solver named Interior Point OPTimizer (IPOPT) \cite{R17}  is  employed  to  solve  the  MPC  problem. The parameters in the MPC-based trajectory planning and tracking control algorithm are selected as $v_c=0.4\rm{m/s}$, $c_v=4$, $H=20$, $H_{\rm{u}}=10$, $Q_i=\textrm{diag}\{1,1,0.01\}$, $R_i=\textrm{diag}\{0.5,0.023\}$, and $S_i=\textrm{diag}\{0.1,0.05\}$. The linear and angular velocities of the AGV are limited by $0 \leq v \le 0.4\rm{m/s}$ and $-0.4\textrm{rad/s} \le w \le 0.4\textrm{rad/s}$, respectively.
We consider the scenario depicted in Fig. \ref{Fig_sim_scenario}. The task of the AGV is to move from point A to point B, and then come back to point C. The original path generated by A* and the MPC-improved path are plotted in Fig. \ref{Fig_MPC_path}. It can be observed that the MPC-improved path enhances the smoothness of the original path.
Fig. \ref{Fig_sim_tra} shows the planned velocity profile. According to the colorbar, one can see that at the turning phase, the velocity is planned to be slow; along the straight lines, the velocity is planned close to the maximum speed.

To evaluate the advantage of the proposed design, three different planning and tracking control schemes are considered: 1) MPC-based trajectory planning + PID tracking control, 2) global planner (A*) + MPC-based tracking control, and 3) the proposed MPC-based trajectory planning and tracking control. To make a fair comparison, these three schemes use the same A* algorithm and their MPC parameters are also the same. The gains for the PID controllers in scheme 1) for the AGV linear and angular velocities are given by the triples (0.065, 0, 0.13) and (0.1, 0.05, 0.2), respectively. Simulation results of the three schemes on the kinematic model of the AGV are depicted in Figs. 6 and 7. The tracking error comparison is summarized in Table \ref{tab:tracking_error}. It can be observed that the proposed approach outperforms the other two approaches in terms of the maximum error $e_{\rm{max}}$, mean error $e_{\rm{mean}}$ and root mean square error (RMSE) $e_{\rm{rmse}}$.

\begin{table}[]
\centering
\caption{Tracking error comparison}
\setlength{\tabcolsep}{3mm}{
\begin{tabular}{c|c c c}
\hline
Planning + Tracking          & $e_{ \rm{max}}$ (m)  & $e_{ \rm{mean}}$ (m) & $e_{\rm{rmse}}$ (m) \\ \hline
MPC + PID        & 0.047 & 0.015 &  0.020\\
A* + MPC        & 0.044 & 0.010 &  0.013\\
MPC + MPC & \textbf{0.028} & \textbf{0.008} &\textbf{0.011}\\ \hline
\end{tabular}\label{tab:tracking_error}}
\end{table}

\begin{figure}[!t]
\begin{center}
  \includegraphics[width=0.38\textwidth]{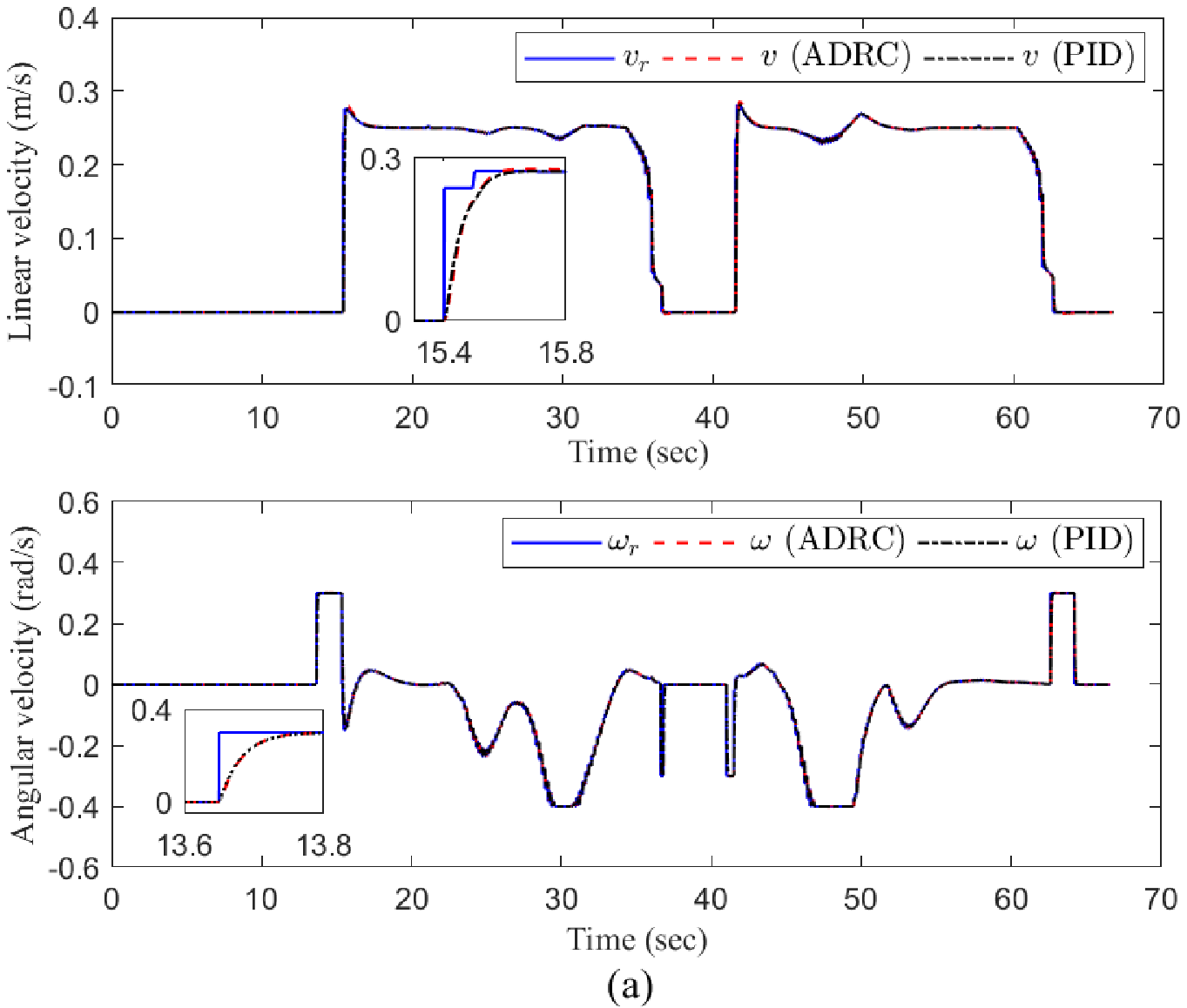}
  \includegraphics[width=0.38\textwidth]{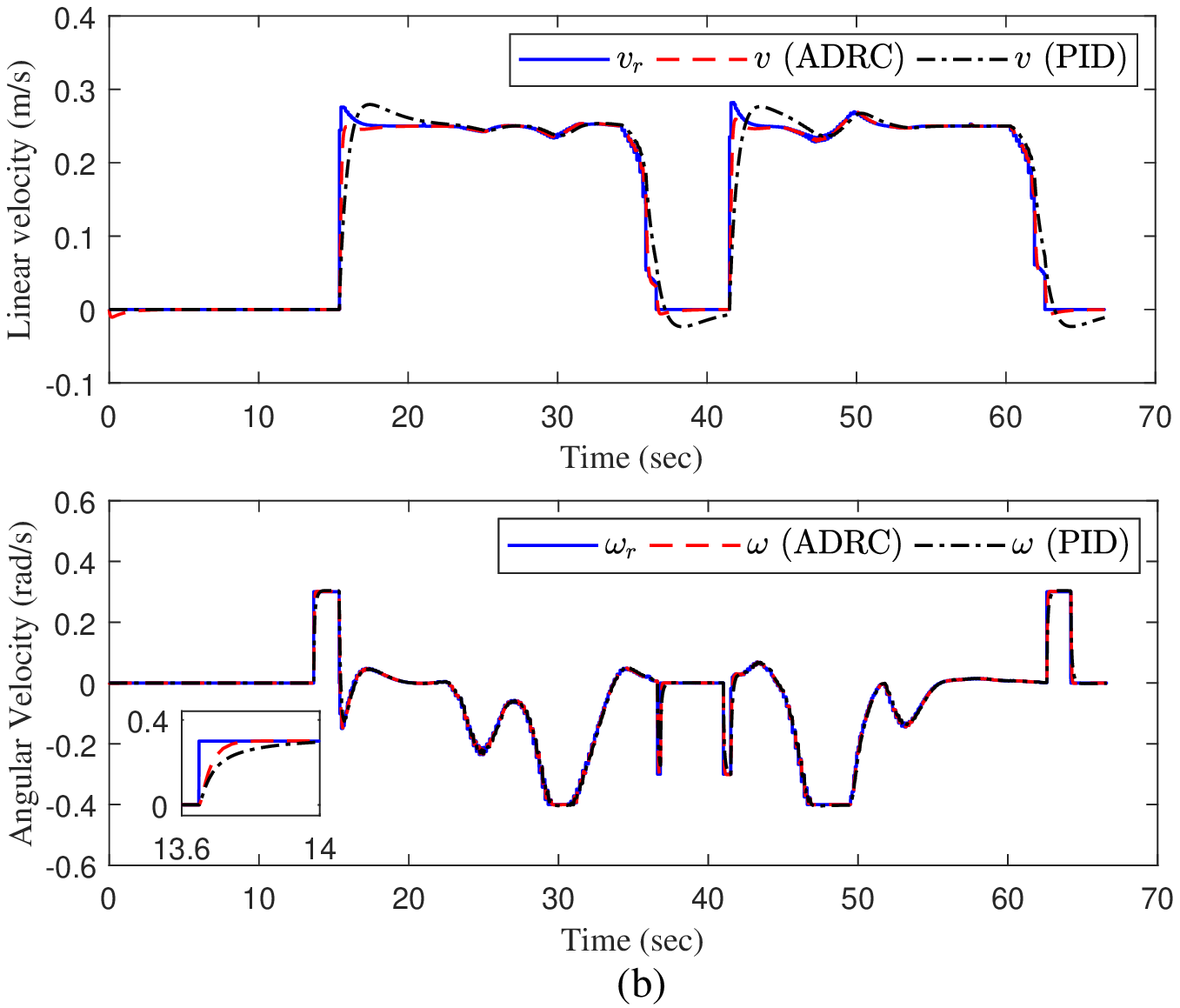}
  \end{center}
    \caption{Performance of the RESO-based controller and PID controller: (a) case 1); (b) case 2).}
    \label{Fig_sim_ADRC_0}
 \end{figure}

\begin{figure}[!t]
\begin{center}
  \includegraphics[width=0.38\textwidth]{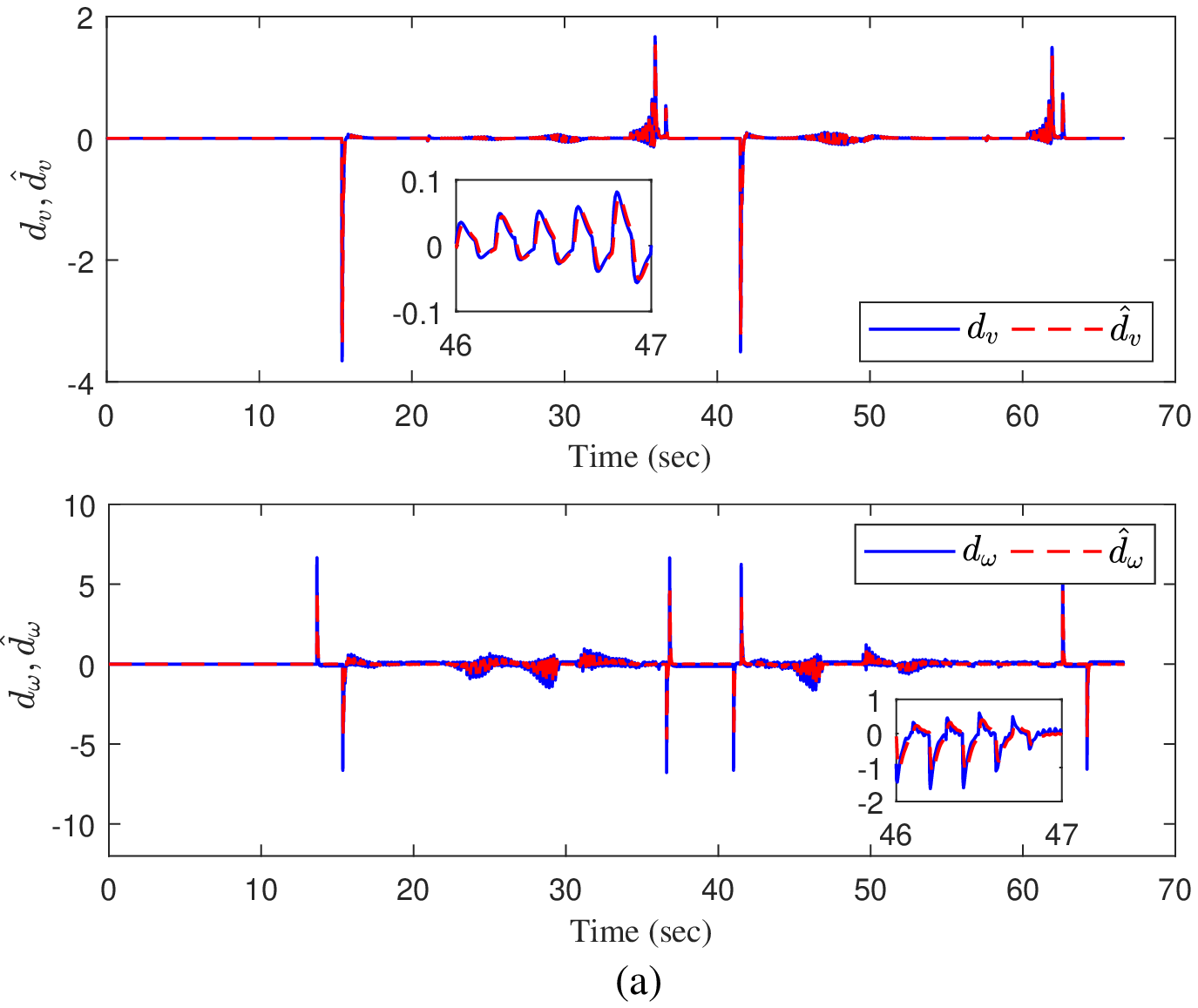}
  \includegraphics[width=0.38\textwidth]{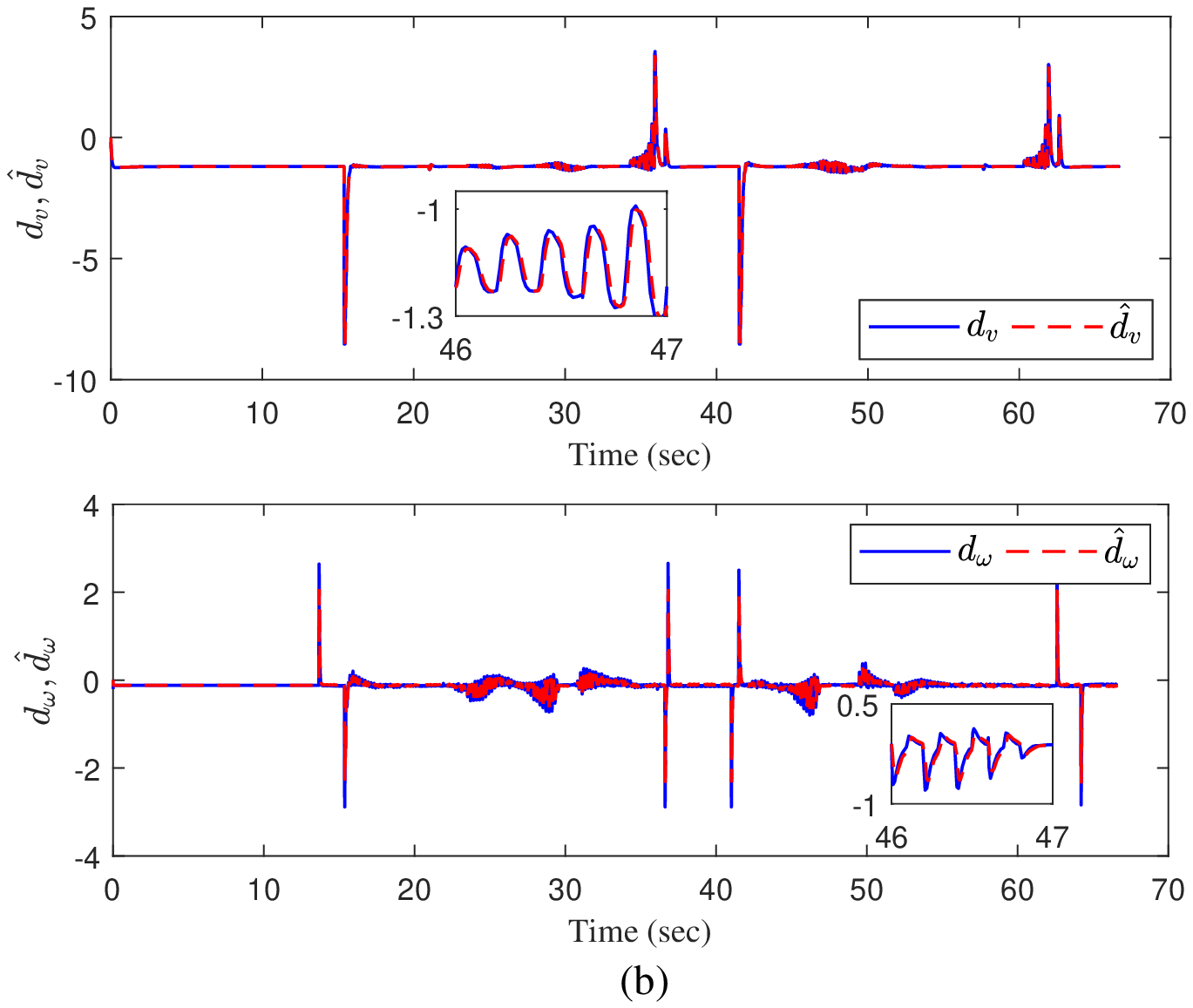}
  \end{center}
  \caption{Performance of the RESO: (a) case 1); (b) case 2).}
 \label{Fig_sim_ADRC_3}
\end{figure}

The parameters of the RESO-based dynamic controller are selected as $\varepsilon=0.01$, $L=1$, $b_0^v=b_0^{\omega}=1$, $K_v=K_{\omega}=-5$, and $M_{v}=M_{\omega}=10$. The linear and angular velocity commands $v_r$ and $\omega_r$ are generated from the MPC-based kinematic controller in Fig. 6c.  For comparison, we also simulate a PID controller whose transfer function is given by $k_P+k_I\frac{1}{s}+k_D\frac{k_N}{1+k_N\frac{1}{s}}$, where $K_N$ is the filter coefficient, and $k_P$, $k_I$, and $k_D$ stand for the proportional, integral, and derivative gains, respectively. The gains of this PID controller are selected as $k_N=100.04$, $k_P=12.74$, $k_I=5.17$, and $k_D=0.88$, which are tuned by the \emph{PID Tuner} function in \emph{Matlab/Simulink} with an overshoot of zero and a rise time of 0.2 seconds. Consider two simulation cases: case 1) without payload and external disturbance; case 2) with payload (triple the weight of the AGV itself) and external disturbance ($f_e=0.2M$ and $\tau_e=0.2I$). Simulation results of the two cases are  shown in Fig. \ref{Fig_sim_ADRC_0}. From this figure, one can see that in case 1) the RESO-based controller and PID controller achieve comparable performance; but in case 2) with large uncertainties, the RESO-based controller performs much better than the PID controller. The main reason for this improvement is that in the proposed RESO-based controller, the total uncertainties are estimated by the observer, and compensated for in the control action in real time. Fig. \ref{Fig_sim_ADRC_3} depicts the performance of the RESO. It can be observed that  the total uncertainties $d_v$ and $d_{\omega}$ are both well-estimated by the RESO in the two cases.

\begin{figure}[!t]
\begin{center}
\includegraphics[width=0.48\textwidth,trim=10 5 5 270,clip]{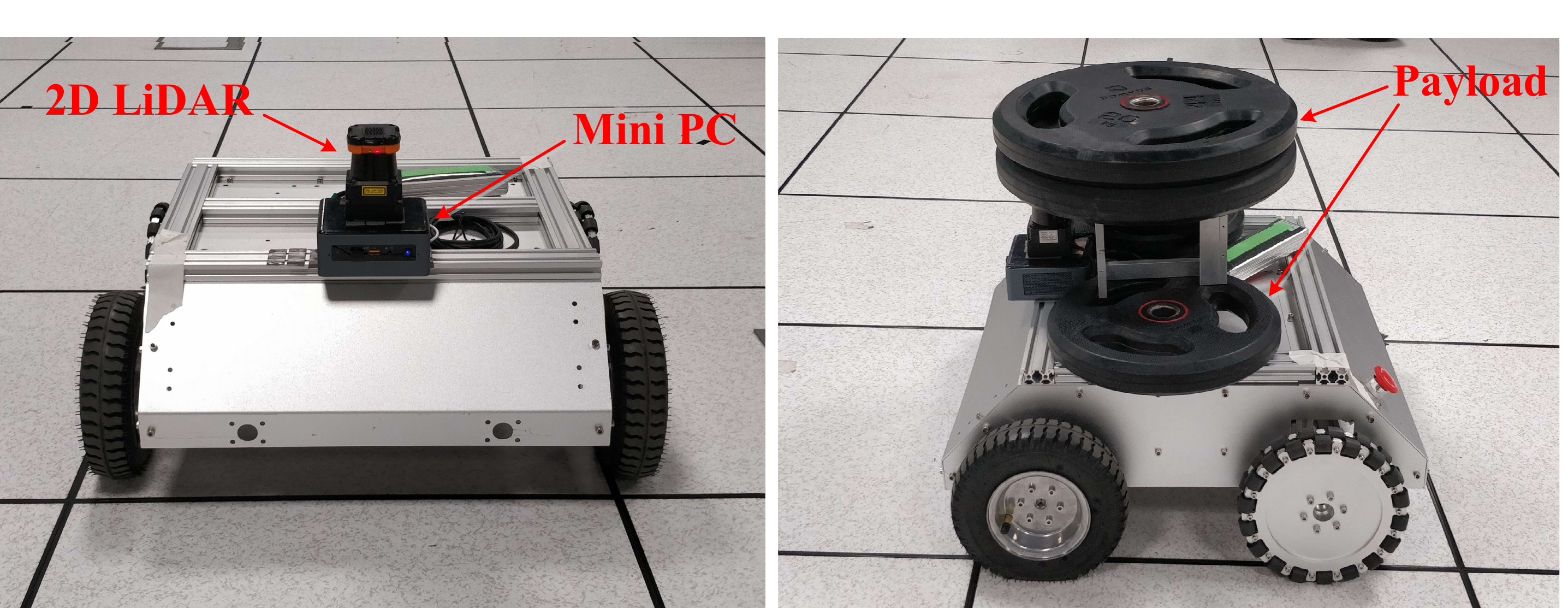}
  \end{center}
  \caption{Experimental platform.}
 \label{Fig_AGV}
\end{figure}

\begin{figure}
     \centering
  \begin{subfigure}
    \centering
    \includegraphics[width=0.38\textwidth]{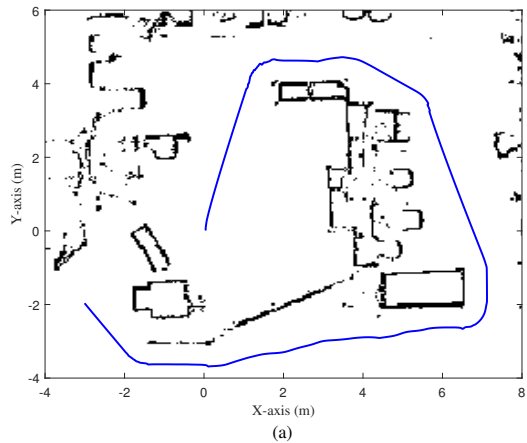}
    \includegraphics[width=0.38\textwidth]{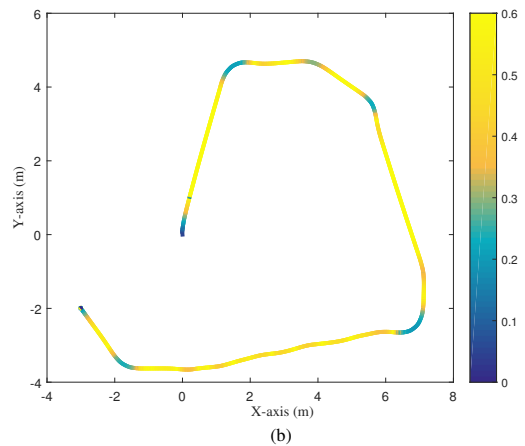}
  \caption{Trajectory planning: (a) map of the environment and the original path; (b) the MPC-improved path and the velocity profile.}
  \label{Fig_exp_map}
  \end{subfigure}
 \end{figure}

\subsection{Experimental Results}

In the experiment, the AGV (see Fig. \ref{Fig_AGV}) is mainly equipped with 1) a 2D LiDAR for localization; 2) a mini PC Intel$^\circledR$ NUC to run the MPC-based trajectory planning and tracking control algorithm; 3)  an ARM$^\circledR$ STM32F103RC MCU to execute the RESO-based dynamic control algorithm; 4) a dual DC motor drive module WSDC2412D to drive the motors. The Adaptive Monte-Carlo Localization (AMCL) algorithm \cite{R18} is implemented to localize the AGV.  The control frequencies of the high-level kinematic control and low-level dynamic control are 20Hz and 100Hz, respectively.

\begin{figure}
    \centering
    \includegraphics[width=0.38\textwidth]{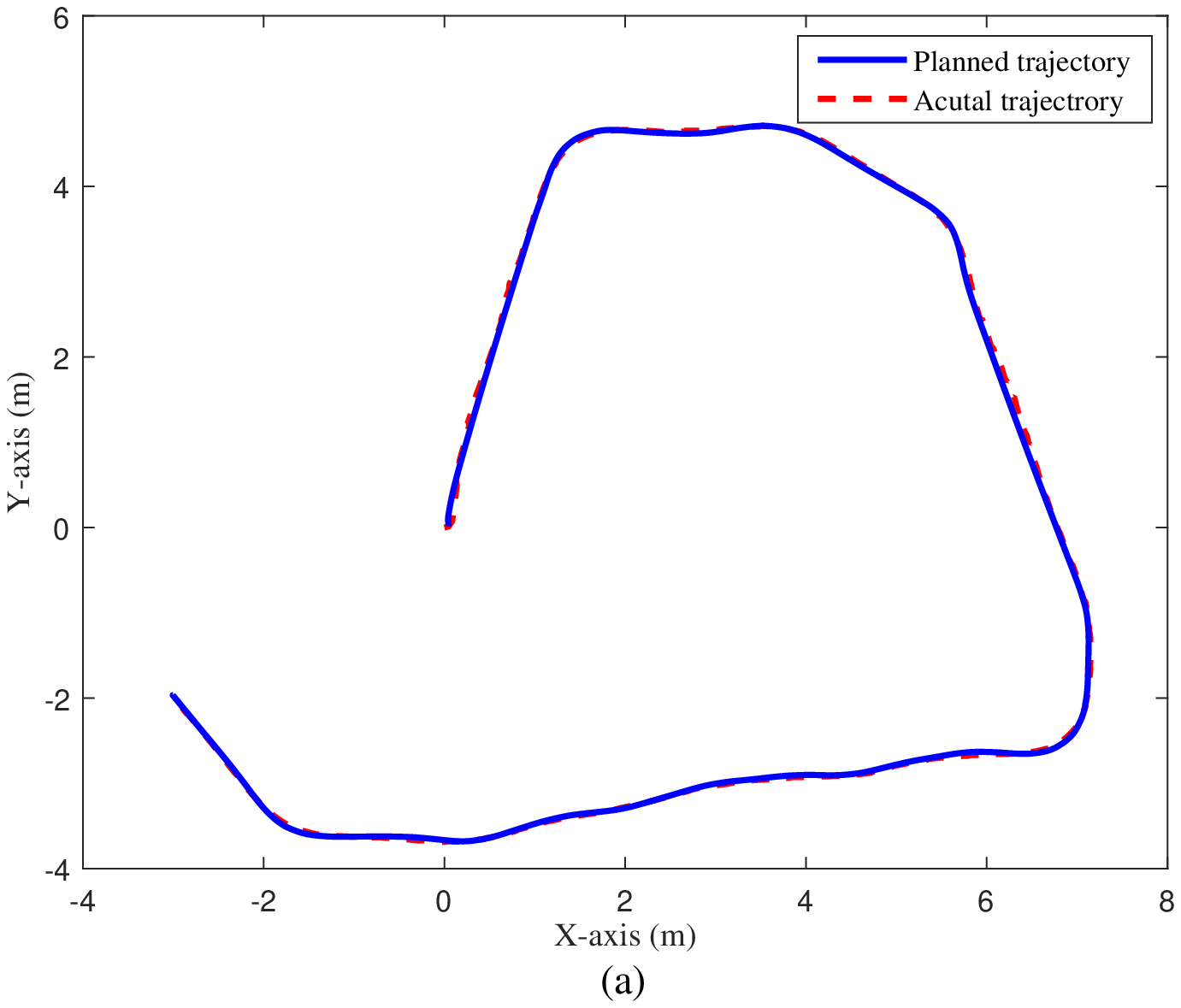}
    \includegraphics[width=0.38\textwidth]{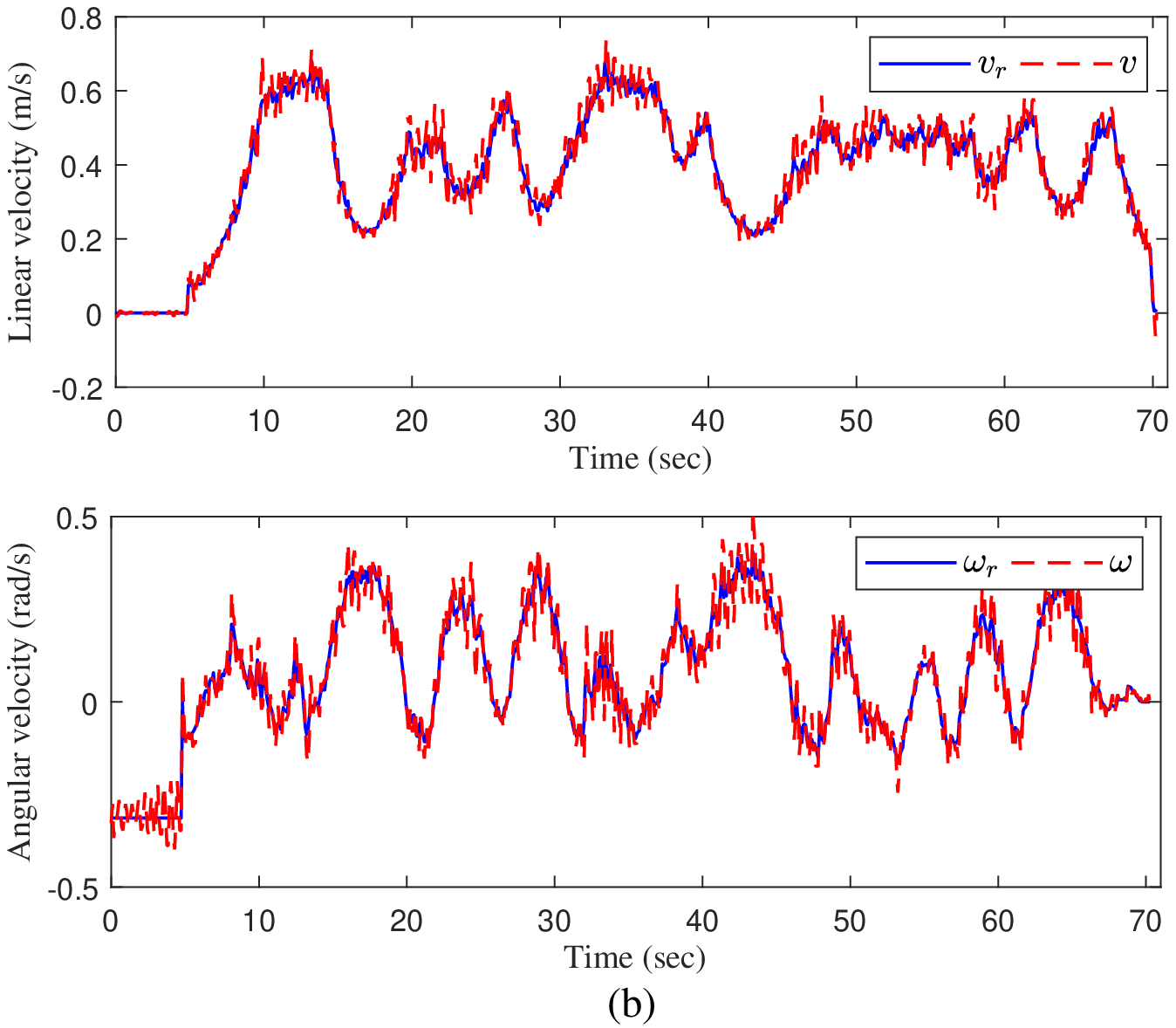}
    \includegraphics[width=0.38\textwidth]{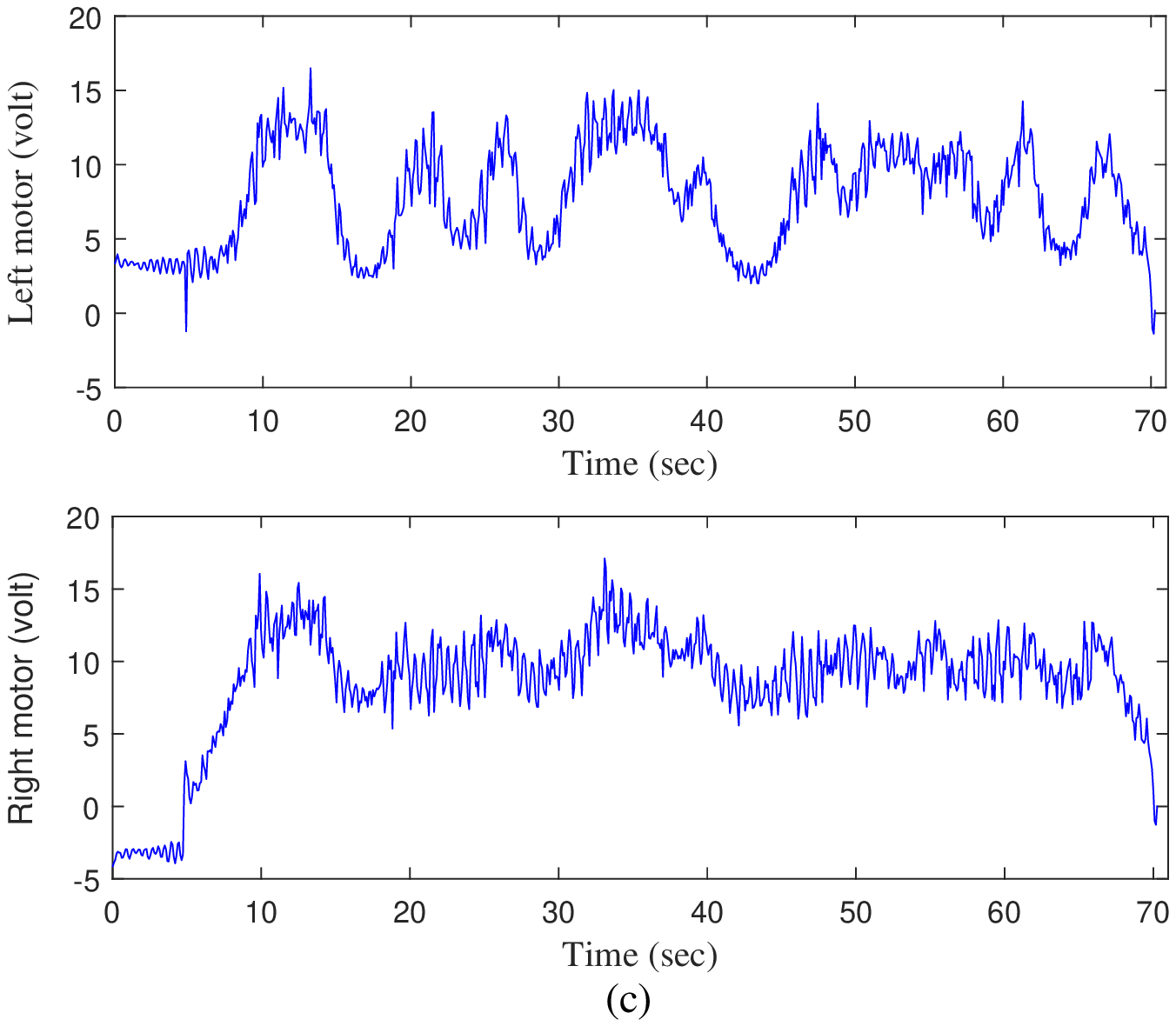}
  \caption{Experimental results with 0 payload: (a) trajectory tracking; (b) response of the RESO-based dynamic controller; (c) motor voltage.}
  \label{Fig_exp_ADRC0}
\end{figure}

\begin{figure}
    \centering
    \includegraphics[width=0.38\textwidth]{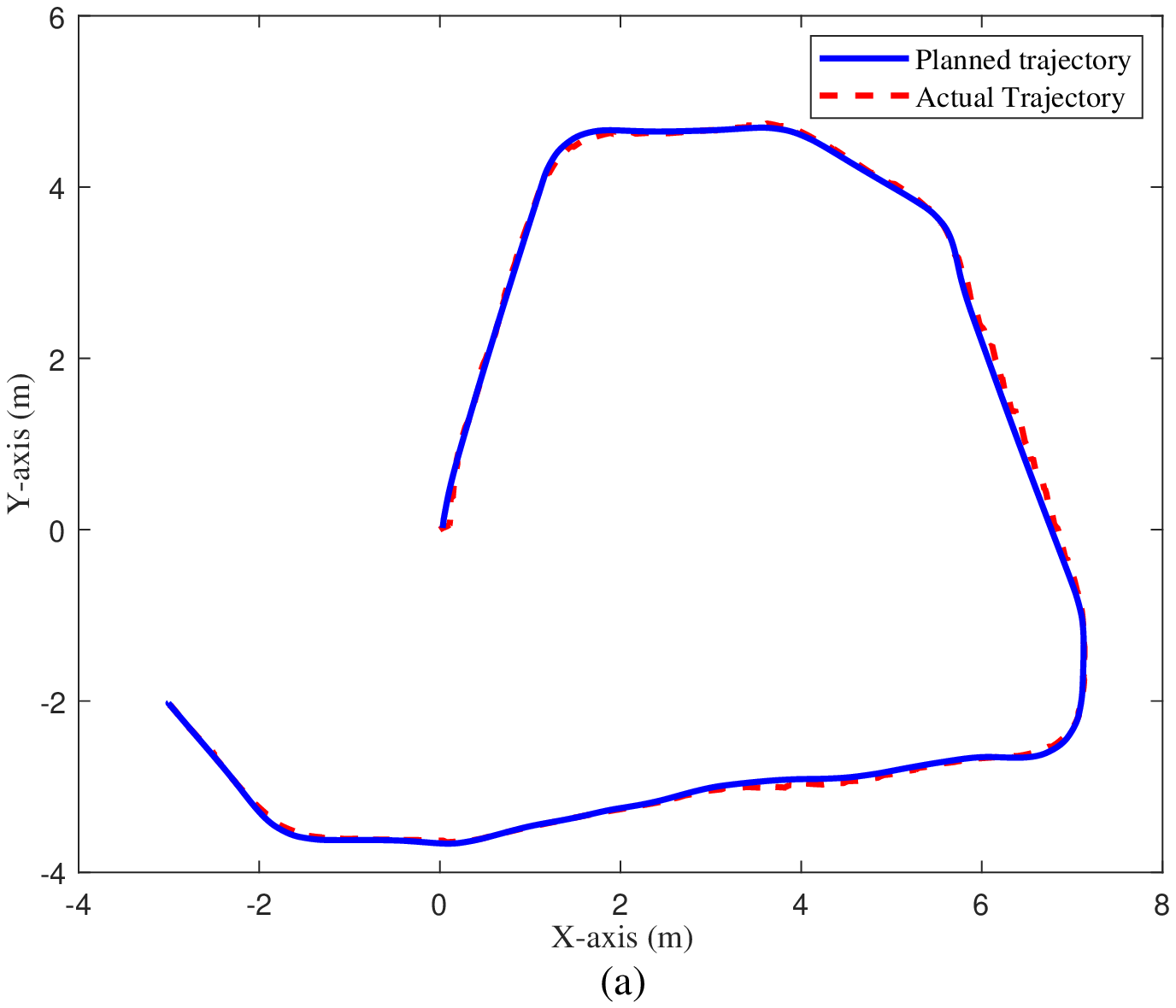}
    \includegraphics[width=0.38\textwidth]{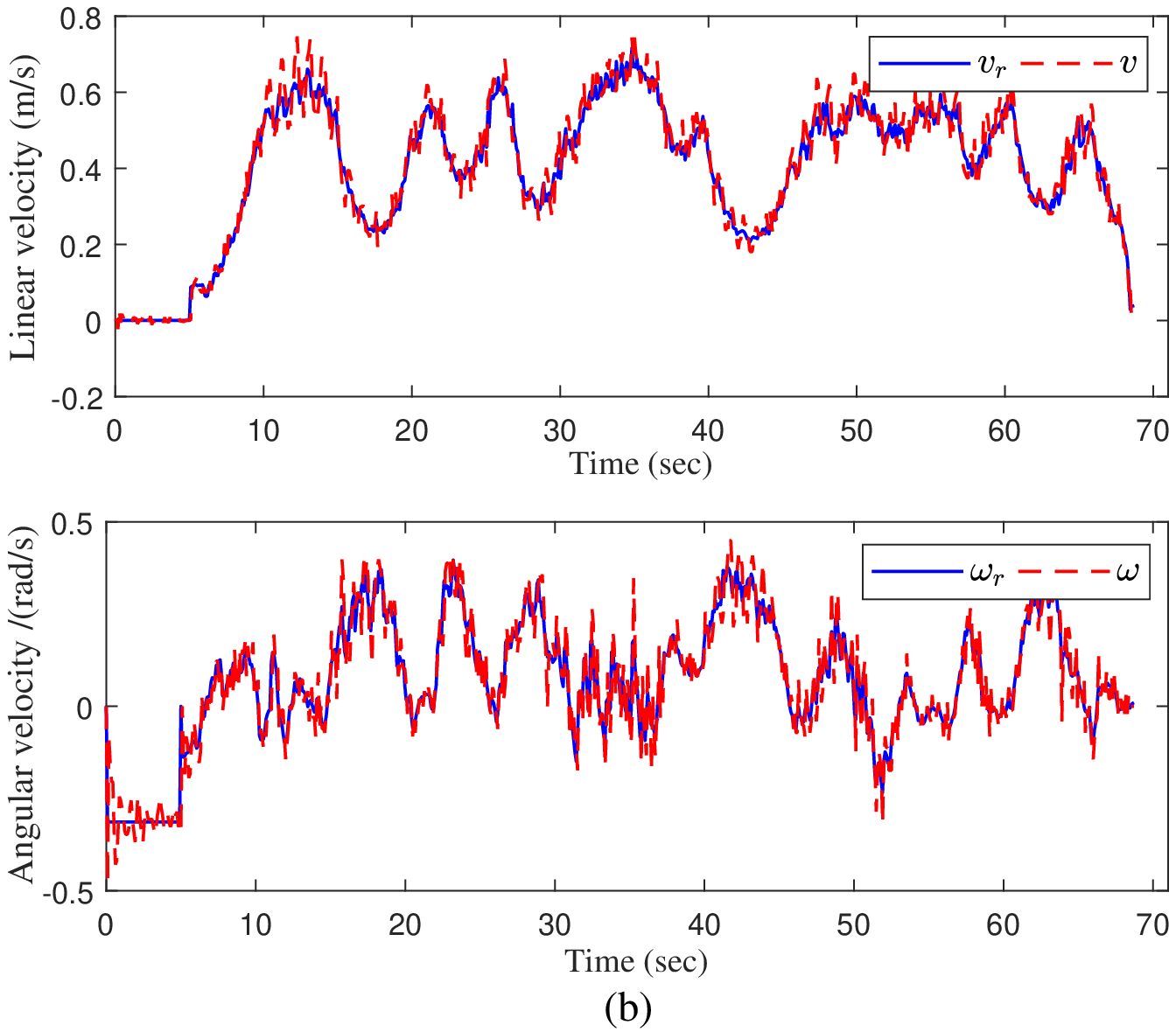}
    \includegraphics[width=0.38\textwidth]{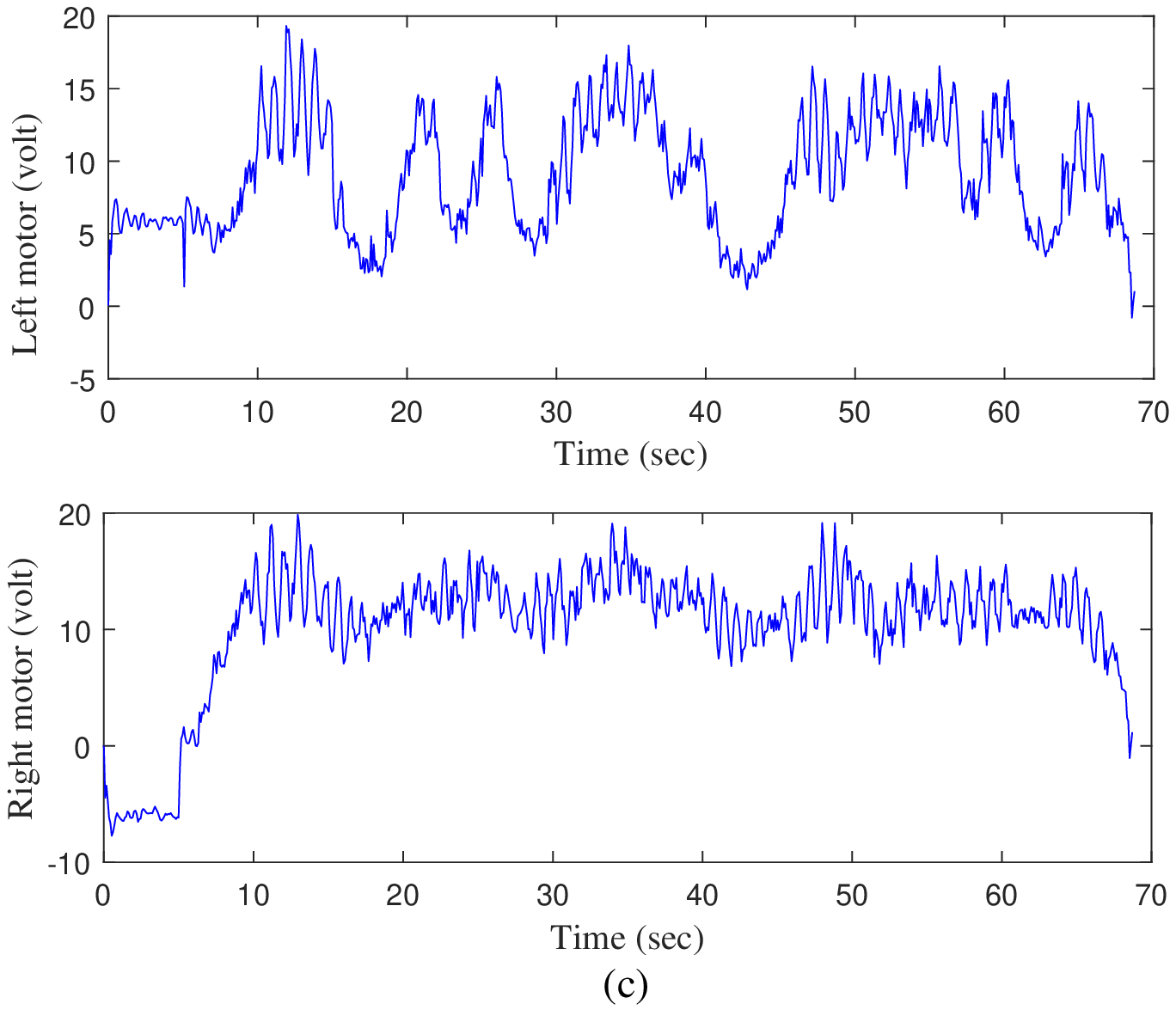}
  \caption{Experimental results with 60kg payload: (a) trajectory tracking; (b) response of the RESO-based dynamic controller; (c) motor voltage.}
  \label{Fig_exp_ADRC60}
\end{figure}

\begin{figure}[!t]
    \centering
    \includegraphics[width=0.38\textwidth]{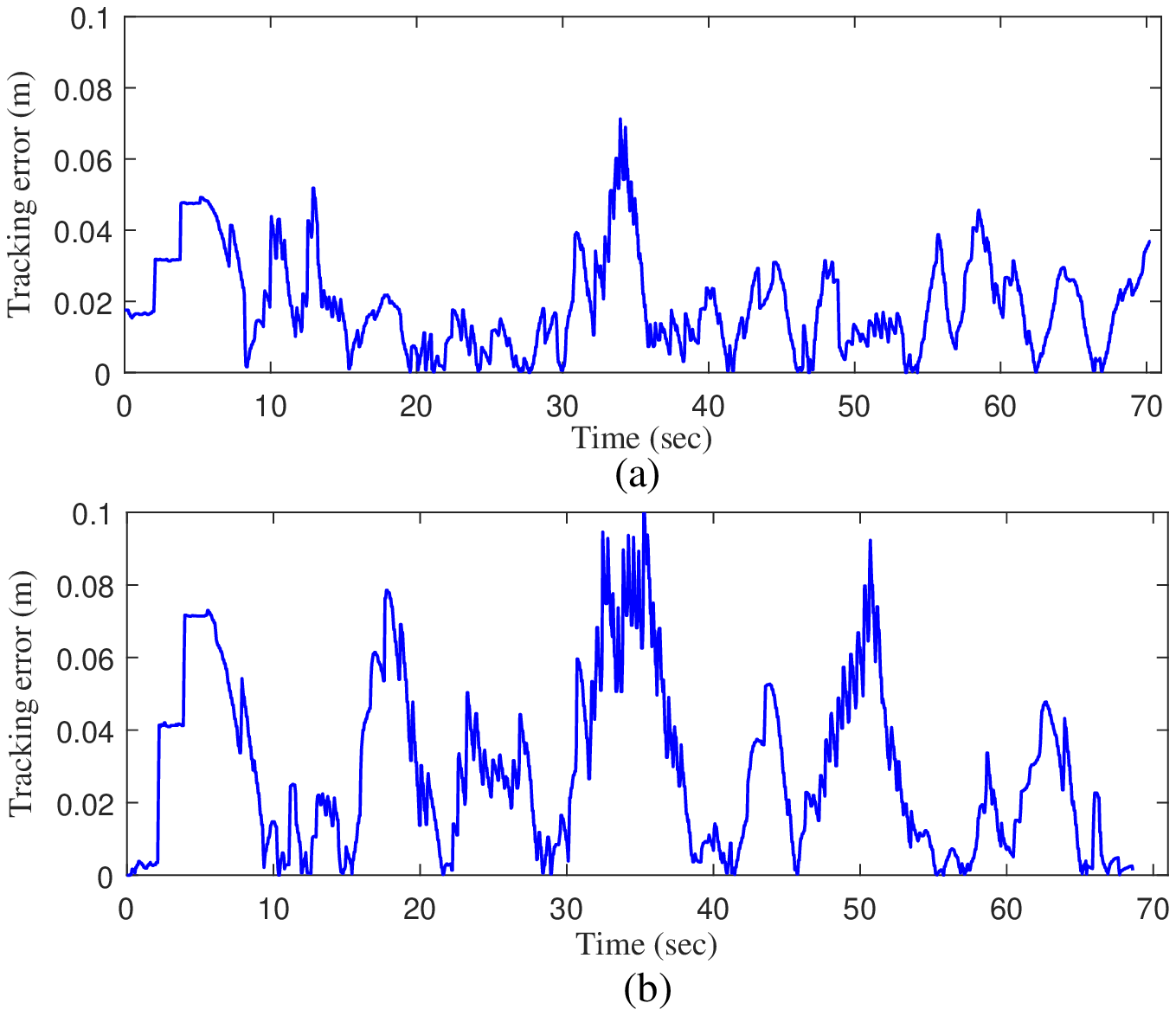}
    \caption{Trajectory tracking errors with different payloads: (a) 0 payload; (b) 60kg payload.}
    \label{Fig_exp_error}
\end{figure}

Experiments are conducted in a warehouse environment. Parameters of the proposed hierarchical controller are inherited from the simulation. Note that in our experiment setting, the actual input of the AGV dynamic model is the motor voltage rather than the torque. The dynamics of the DC motor can be approximated by a first-order inertial system with a small time constant \cite{R3}. Since the RESO-based controller is capable of handling large uncertainties and only relies on the sign of the control gain, in the implementation stage, the commands of the voltage of the DC motors are also given by (\ref{eq22}). Fig. 11a illustrates the map of the environment and the original path generated by the A$^*$ algorithm.  Fig. 11b shows the MPC-improved path and the reference velocity. The maximal velocity of the AGV in the experiments is set as 0.6m/s. Figs. \ref{Fig_exp_ADRC0} and \ref{Fig_exp_ADRC60} depict the experimental results with 0 payload and 60kg payload, respectively. Note that 60kg payload is about double the weight of the AGV itself, which indicates large uncertainty corresponding to the AGV dynamics. Such uncertainty will increase the burden of the controller, and has stubborn effects on the overall control performance. However, from Figs. \ref{Fig_exp_ADRC0} and \ref{Fig_exp_ADRC60}, one can observe that the AGV moves along the planned trajectory accurately, the commands generated by the MPC-based kinematic controller are well-tracked by the RESO-based dynamic controller, and the voltages of the two motors are also acceptable. Fig. \ref{Fig_exp_error} depicts the trajectory tracking error with 0 payload and 60kg payload. With 0 payload, the maximal tracking error is 0.071m, the average tracking error is 0.019m; with 60kg payload, the maximal tracking error is 0.105m, the average tracking error is 0.029m. The experimental video is available at \url{https://youtu.be/SxdO9YXbiZs}.

\section{Conclusion}

A hierarchical control scheme is proposed for AGVs with large uncertainties. The MPC-based trajectory planning and tracking control at the high level provides satisfactory trajectory and accurate kinematic tracking performance, while the RESO-based dynamic control at the low level handles the large uncertainties. The proposed hierarchical control scheme needs little information of the AGV dynamics, and is simple for implementation. Experimental results for an AGV with different payloads verified  the effectiveness of the proposed approach in handling large uncertainties. In future studies, the focus will be on the extension of the hierarchical scheme to multiple-AGVs in complex manufacturing environment.  The uncertainties caused by the environment and communication will be considered.

\vspace{1cm}
\appendix

We need a lemma before giving the proof of Theorem 1. Define the tracking error $e(t)=\eta(t)-\varrho(t)$, and a Lyapunov function candidate $V(e(t))=\frac{1}{2}e^2(t)$.  Denote $\tau_0=V(e(0))+1$, and define two compact sets:
\begin{align*}
\Omega_0=&\{e(t)\in\mathbb{R}; V(e(t))\leq \tau_0\}, \\
\Omega_1=&\{e(t)\in\mathbb{R}; V(e(t))\leq \tau_0+1\}.
\end{align*}
Note that $\Omega_0\subseteq \Omega_1$ and $e(0)$ is an internal point of $\Omega_0$.  The following lemma shows that for sufficiently small $\varepsilon$, $e(t)\in\Omega_1$, $\forall t\in[0, \infty)$.

\emph{Lemma 1:} Consider the closed-loop system formed of (\ref{eq2}), (\ref{eq4}), (\ref{eq5}), and (\ref{eq7}). Suppose Assumptions A1 to A3 are satisfied,  and the initial conditions $\eta(0)$ and $\varsigma(0)$ are bounded. Then there exists $\varepsilon^*>0$ such that for any $\varepsilon\in(0, \varepsilon^*)$, $e(t)\in\Omega_1$, $\forall t\in[0, \infty)$.

\emph{Proof of Lemma 1}: Since $e(0)$ is an internal point of $\Omega_0$, and the control $u(t)$ is bounded, there exits an $\varepsilon$-independent $t_0>0$ such that $e(t)\in \Omega_0$, $\forall t\in[0, t_0]$. Lemma 1 will be proved by contradiction.  Suppose Lemma 1 is false, then there exist $t_2>t_1>t_0$ such that
\begin{equation}\label{eq15}
 \left\{
 \begin{aligned}
 & V(e(t_1))=\tau_0,  \\
 & V(e(t_2))=\tau_0+1,   \\
 &  \tau_0 \leq V(e(t))\leq \tau_0+1, ~t\in[t_1, t_2],  \\
 & V(e(t))\leq \tau_0+1,  ~t\in [0, t_2].
 \end{aligned}
 \right.
\end{equation}

Consider the RESO estimation error $\delta(t)=\xi(t)-\hat{\xi}(t)$. By (\ref{eq3})-(\ref{eq5}), the dynamics of $\delta(t)$ can be formulated as\footnote{For notation simplicity, we omit the time symbol $t$ occasionally.}
\begin{align}\label{eq12}
	\dot{\delta}= & \dot{\xi}-\dot{\hat{\xi}} \nonumber \\
	 = & \dot{\xi}-\frac{1}{\varepsilon}L\left(f(\eta, \varpi)+b(\eta, \varpi)u-\frac{1}{\varepsilon}L(\eta-\varsigma)-b_0(\eta)u\right)  \nonumber\\
	 	 = & \dot{\xi}(t)-\frac{1}{\varepsilon}L\left(f(\eta, \varpi)+(b(\eta, \varpi)-b_0(\eta))u-\hat{\xi}\right) \nonumber\\
	 	 = &\dot{\xi}-\frac{1}{\varepsilon}L(\xi-\hat{\xi}) \nonumber\\
	 = &  \dot{\xi}-\frac{1}{\varepsilon}L\delta.
\end{align}
The differentiation of the extended state $\xi(t)$ can be computed as
\begin{align}\label{eq10a}
\dot{\xi} = & \frac{\textrm{d}f}{\textrm{d}t}(\eta, \varpi)+\left(\frac{\textrm{d}b}{\textrm{d}t}(\eta, \varpi)-\frac{\textrm{d} b_0}{\textrm{d}t}(\eta)\right)u \nonumber \\
                        & +(b(\eta, \varpi)-b_0(\eta))\dot{u},
 \end{align}
 where
 \begin{align}
 \frac{\textrm{d}f}{\textrm{d}t}(\eta, \varpi)
                        =&  \left(f(\eta, \varpi)+b(\eta, \varpi)u\right)\frac{\partial f}{\partial \eta}(\eta, \varpi) +\dot{\varpi}\frac{\partial f}{\partial \varpi}(\eta, \varpi), \nonumber \\
 \frac{\textrm{d}b}{\textrm{d}t}(\eta, \varpi)
                        =&  \left(f(\eta, \varpi)+b(\eta, \varpi)u\right)\frac{\partial b}{\partial \eta}(\eta, \varpi)+\dot{\varpi}\frac{\partial b}{\partial \varpi}(\eta, \varpi),  \nonumber \\
  \frac{\textrm{d}b_0}{\textrm{d}t}(\eta)
                        = &   \left(f(\eta, \varpi)+b(\eta, \varpi)u\right) \frac{\partial b_0}{\partial \eta}(\eta). \nonumber
\end{align}
By (\ref{eq5}) and (\ref{eq6}), the control $u$ can be expressed as
\begin{equation*}\label{eq23}
	u=M_u\textrm{sat}_{\varepsilon}\left(\frac{K(\eta-\varrho)-\hat{\xi}+\dot{\varrho}}{M_ub_0(\eta)}\right).
\end{equation*}
It follows that $\dot{u}$ can be computed as
\begin{align}\label{eq24}
 \dot{u} = & \frac{\textrm{dsat}_{\varepsilon}(\cdot)}{\textrm{d}(\cdot)} \frac{1}{b_0^2(\eta)}\left(\left(K(\dot{\eta}-\dot{\varrho})-\dot{\hat{\xi}}+\ddot{\varrho}\right)  b_0(\eta)\right. \nonumber\\
 & \left. -\left(K(\eta-\varrho)-\hat{\xi}+\dot{\varrho}\right)\frac{\textrm{d}b_0}{\textrm{d}t}(\eta)\right) \nonumber  \\
 = &   \frac{\textrm{dsat}_{\varepsilon}(\cdot)}{\textrm{d}(\cdot)} \frac{1}{b_0(\eta)} \left(K\left(f(\eta, \varpi)+b(\eta,\varpi)u-\dot{\varrho}\right)+\ddot{\varrho}\right) \nonumber \\
 & -\frac{\textrm{dsat}_{\varepsilon}(\cdot)}{\textrm{d}(\cdot)} \frac{1}{b_0^2(\eta)}\frac{\textrm{d}b_0}{\textrm{d}t}(\eta)\left(Ke-\xi+\dot{\varrho}\right) \nonumber \\
  & -\frac{\textrm{dsat}_{\varepsilon}(\cdot)}{\textrm{d}(\cdot)} \frac{1}{b_0^2(\eta)}\frac{\textrm{d}b_0}{\textrm{d}t}(\eta)\delta -\frac{\textrm{dsat}_{\varepsilon}(\cdot)}{\textrm{d}(\cdot)} \frac{1}{b_0(\eta)}\frac{1}{\varepsilon}L\delta.
 \end{align}
By (\ref{eq10a}) and (\ref{eq24}), one has
\begin{align}\label{eq10}
\dot{\xi} = & \frac{\textrm{d}f}{\textrm{d}t}(\eta, \varpi)+\left(\frac{\textrm{d}b}{\textrm{d}t}(\eta, \varpi)-\frac{\textrm{d} b_0}{\textrm{d}t}(\eta)\right)u \nonumber \\
 &  + \frac{\textrm{dsat}_{\varepsilon}(\cdot)}{\textrm{d}(\cdot)} \frac{b(\eta, \varpi)-b_0(\eta)}{b_0(\eta)} \nonumber \\
 & \times \left(K\left(f(\eta, \varpi)+b(\eta,\varpi)u-\dot{\varrho}\right)+\ddot{\varrho}\right) \nonumber \\
 & -\frac{\textrm{dsat}_{\varepsilon}(\cdot)}{\textrm{d}(\cdot)} \frac{b(\eta, \varpi)-b_0(\eta)}{b_0^2(\eta)}\frac{\textrm{d}b_0}{\textrm{d}t}(\eta)\left(Ke-\xi+\dot{\varrho}\right) \nonumber \\
 & -\frac{\textrm{dsat}_{\varepsilon}(\cdot)}{\textrm{d}(\cdot)} \frac{b(\eta, \varpi)-b_0(\eta)}{b_0^2(\eta)}\frac{\textrm{d}b_0}{\textrm{d}t}(\eta)\delta \nonumber \\
 & -\frac{\textrm{dsat}_{\varepsilon}(\cdot)}{\textrm{d}(\cdot)} \frac{b(\eta, \varpi)-b_0(\eta)}{b_0(\eta)}\frac{1}{\varepsilon}L\delta.
 \end{align}
 By Assumptions A1, A2,  and the boundedness of  $e(t)$ in the time interval $[0, t_2]$, one can conclude from (\ref{eq10}) that there exist $\varepsilon$-independent positive constants $C_0$ and $C_1$ such that
\begin{equation}\label{eq11}
\delta\dot{\xi}\leq C_0|\delta|+C_1\delta^2-\frac{1}{\varepsilon}L\Delta\delta^2,
\end{equation}
where
\begin{equation*}
    \Delta=\frac{\textrm{dsat}_{\varepsilon}(\cdot)}{\textrm{d}(\cdot)} \frac{b(\eta, \varpi)-b_0(\eta)}{b_0(\eta)}.
\end{equation*}

Consider the Lyapunov function candidate $W(\delta)=\frac{1}{2}\delta^2$. It follows from (\ref{eq12}) and (\ref{eq11}) that
\begin{align}\label{eq13}
   \dot{W}(\delta)= & \delta\left(\dot{\xi}-\frac{1}{\varepsilon}L\delta\right) \nonumber \\
   \leq & -\frac{1}{\varepsilon}L(1+\Delta)\delta^2+C_1\delta^2+C_0|\delta|.
\end{align}
Note that the term $1+\Delta$ satisfies
	\begin{align}\label{eq25}
	  1+\Delta=&1+\frac{\textrm{dsat}_{\varepsilon}(\cdot)}{\textrm{d}(\cdot)} \frac{b(\eta, \varpi)-b_0(\eta)}{b_0(\eta)} \nonumber \\
	  =&\frac{1}{b_0(\eta)}\left[\frac{\textrm{dsat}_{\varepsilon}(\cdot)}{\textrm{d}(\cdot)}b(\eta,\varpi)+\left(1-\frac{\textrm{dsat}_{\varepsilon}(\cdot)}{\textrm{d}(\cdot)}\right)b_0(\eta)\right].
	\end{align}
Since $b(\eta,\varpi)$ and $b_0(\eta)$  have the same sign,  and $0\leq\frac{\textrm{dsat}_{\varepsilon}(\cdot)}{\textrm{d}(\cdot)}\leq 1$,  (\ref{eq25}) guarantees that $1+\Delta>0$. It then follows from (\ref{eq13}) that  there exists sufficiently small $\varepsilon_1>0$ such that for any $\varepsilon\in(0,\varepsilon_1)$ and $t\in[t_0,t_2]$, $|\delta(t)|=O(\varepsilon)<-K\sqrt{2\tau_0}$.
	
Next, we show that the function $\textrm{sat}_{\varepsilon}(\cdot)$ in (\ref{eq7}) is out of saturation in the time interval $[t_0, t_2]$. By the convergence of the RESO, one has
\begin{equation}
    \hat{\xi}=f(\eta,\varpi)+(b(\eta,\varpi)-b_0(\eta))u+O(\varepsilon).
\end{equation}
Therefore, up to an $O(\varepsilon)$ error, $\psi(\eta,\varrho,\hat{\xi})$ satisfies the equation
\begin{align}
\psi+\frac{b(\eta,\varpi)-b_0(\eta)}{b_0(\eta)}M_u\textrm{sat}_{\varepsilon}\left(\frac{\psi}{M_u}\right) \quad \nonumber \\
=\frac{K(\eta-\varrho)-f(\eta,\varpi)+\dot{\varrho}}{b_0(\eta)}.
\end{align}
This equation has a unique solution since $1+\Delta>0$. Note that
the saturation bound $M_u$ satisfies (\ref{bound}). It can be obtained by direct substitution that the unique solution is
\begin{equation}
    \psi^*=\frac{K(\eta-\varrho)-f(\eta,\varpi)+\dot{\varrho}}{b(\eta,\varpi)}.
\end{equation}
Since the saturation bound $M_u$ is selected to satisfy (\ref{bound}), one can conclude that for sufficiently small $\varepsilon$, $\psi(\eta,\varrho,\hat{\xi})$ will be in the linear region of the saturation function in the time interval $[t_0,t_2]$.  It follows that the time derivative of $V(e)$ can be computed as
\begin{align}\label{eq14}
   \dot{V}(e)= &  e(\dot{\eta}-\dot{\varrho}) \nonumber \\
  = & e\left(\xi+Ke-\hat{\xi}\right) \nonumber \\
  \leq & Ke^2+|e|\cdot|\delta|.
\end{align}
By (\ref{eq15}),  in the time interval $[t_1,t_2]$,  $\sqrt{2\tau_0}\leq |e|\leq \sqrt{2\tau_0+2}$. This together with the relation $|\delta|<-K\sqrt{2\tau_0}$ yields
\begin{equation}
    \dot{V}(e)\leq K|e|\left(|e|+\frac{1}{K}|\delta|\right)<0, ~t\in[t_1, t_2],
\end{equation}
which, contradicts (\ref{eq15}). Thus there exists $\varepsilon^*>0$ such that for any $\varepsilon\in(0, \varepsilon^*)$, $e(t)\in\Omega_1$, $\forall t\in[0, \infty)$. This completes the proof of Lemma 1. \IEEEQED

Based on Lemma 1, we are ready to state the proof of Theorem 1.

\emph{Proof of Theorem 1:} By Lemma 1, $e(t)\in\Omega_1$ for any $\varepsilon\in(0,\varepsilon^*)$ and $t\in[0,\infty)$. It follows that (\ref{eq11}) and (\ref{eq13}) hold for any $\varepsilon\in(0,\varepsilon^*)$ and $t\in[0,\infty)$.  By (\ref{eq13}), one has that for any $\sigma>0$ and $T>0$, there exists $\varepsilon^1\in (0,\varepsilon^*]$ such that for any $\varepsilon\in(0,\varepsilon^1)$, $|\xi(t)-\hat{\xi}(t)|\leq \sigma$, $t\in [T,\infty)$. By  (\ref{eq14}), one can conclude that there exists $\varepsilon^{\dag}\in(0,\varepsilon^1]$ such that (\ref{eq9}) holds. This completes the proof of Theorem 1. \IEEEQED
\vspace{6pt}

\balance

\end{document}